\PassOptionsToPackage{square,sort&compress,comma,numbers}{natbib}
\documentclass[final,5p,times,twocolumn]{elsarticle}
\usepackage{subfigure}
\usepackage{amsmath}
\usepackage{bm}
\usepackage{graphicx}
\usepackage{epstopdf}
\usepackage{hyperref}
\usepackage{setspace}
\usepackage{lineno}
\usepackage{geometry}

\usepackage{amssymb}

\journal{Robotics and Autonomous Systems}

\begin{document}

\onehalfspacing


\begin{frontmatter}



\title{ATDM:An Anthropomorphic Aerial Tendon-driven Manipulator with Low-Inertia and High-Stiffness}



\author[label1]{Quman XU}
\author[label1,label2]{Zhan Li\corref{cor1}}
\author[label1]{Hai Li}
\author[label3]{Xinghu Yu}
\author[label1]{Yipeng Yang}

\cortext[cor1]{Corresponding author. \href{mailto:zhanli@hit.edu.cn}{E-mail address: zhanli@hit.edu.cn} (Z. Li).}

\affiliation[label1]{organization={Research Institute of Intelligent Control and Systems, Harbin Institute of Technology},
            addressline={},
            city={Harbin},
            postcode={150001},
            state={Heilongjiang},
            country={China}}

\affiliation[label2]{organization={Peng Cheng Laboratory},
            addressline={},
            city={Shenzhen}, 
            postcode={518000},
            state={Guangdong},
            country={China}}

\affiliation[label3]{organization={Ningbo Institute of Intelligent Equipment Technology Co., Ltd.}, 
            addressline={}, 
            city={Ningbo},
            postcode={315201},
            state={Zhejiang},
            country={China}
}

\begin{abstract}

Aerial Manipulator Systems (AMS) have garnered significant interest for their utility in aerial operations. Nonetheless, challenges related to the manipulator's limited stiffness and the coupling disturbance  with manipulator movement persist. This paper introduces the Aerial Tendon-Driven Manipulator (ATDM), an innovative AMS that integrates a hexrotor Unmanned Aerial Vehicle (UAV) with a 4-degree-of-freedom (4-DOF) anthropomorphic tendon-driven manipulator. The design of the manipulator is anatomically inspired, emulating the human arm anatomy from the shoulder joint downward. To enhance the structural integrity and performance, finite element topology optimization and lattice optimization are employed on the links to replicate the radially graded structure characteristic of bone, this approach effectively reduces weight and inertia while simultaneously maximizing stiffness. A novel tensioning mechanism with adjustable tension is introduced to address cable relaxation, and a Tension-amplification tendon mechanism is implemented to increase the manipulator's overall stiffness and output. The paper presents a kinematic model based on virtual coupled joints, a comprehensive workspace analysis, and detailed calculations of output torques and stiffness for individual arm joints.

The prototype arm has a total weight of 2.7 kg, with the end effector contributing only 0.818 kg. By positioning all actuators at the base, coupling disturbance are minimized. The paper includes a detailed mechanical design and validates the system's performance through semi-physical multi-body dynamics simulations, confirming the efficacy of the proposed design.

\end{abstract}





\end{frontmatter}


\section{Introduction}

UAVs, with their immense research value, are forecasted to reach a global market size of US\$ 17,520 million by 2028\footnote{GlobeNewswire. URL: \href{https://www.globenewswire.com/news-release/2022/06/17/2464627/0/en/Drone-Market-Size-Share-2022-2029-Insights-and-Forecast-Research-Growth-Rate-Demands-Trends-Key-Players-Geographical-Segmentation-Sales-Price-Revenue-and-Gross-Margin-Analysis-Mark}{https://www.globenewswire.com}. Accessed on: [2023/10/31].}. Known for their ease of operation and high agility and maneuverability in three-dimensional space, UAVs have applications in areas such as aerial photography, mapping, and agriculture\cite{song2022anti,ramos2022hybrid,song2023feedforward}. Historically, their use was primarily restricted to passive monitoring tasks. Inspired by avian flight, researchers have integrated multi-DOF mechanical manipulators or enhancement mechanisms to flying platforms, leading to the development of AMS. This allows UAVs to actively interact with their environment\cite{BONYANKHAMSEH2018221}, attracting significant interest from both the academic and industrial sectors.

Currently, AMS are used for tasks that pose inconveniences or hazards to humans, such as aerial grasping~\cite{zhang2018grasp}, aerial manipulation~\cite{acosta2020accurate,xu2023aerial,dong2021centimeter,sun2021switchable}, contact inspection~\cite{suarez2020aerial,zeng2023autonomous}, bridge maintenance~\cite{ikeda2019stable}, and environmental monitoring~\cite{roderick2021bird}. Aside from underactuated compliant mechanisms like SNAG~\cite{roderick2021bird} and single-link or grippers~\cite{mclaren2019passive,bodieomnidirectional}, the majority of AMS prototypes feature a serial manipulator configuration comprising an upper arm and a forearm. The choice in the number of joints and the kinematic arrangement is primarily influenced by the operation's needed dexterity, aiming to reduce the number of actuators and consequently the overall weight~\cite{ollero2021past}.

The control of AMS is challenging due to their intrinsic coupling, non-linear dynamics, and uncertainties \cite{khamseh2019unscented, samadikhoshkho2022vision}. Addressing these challenges, Cao et al. \cite{cao2022adaptive} employ a radial basis function neural network (RBFNN) for real-time dynamic coupling compensation between UAVs and manipulators; Emami et al. \cite{emami2021simultaneous} propose a multi-stage disturbance observer-based model predictive control(MPC) for aerial grasping and trajectory tracking; and  Chen et al.\cite{chen2022adaptive} explore an adaptive sliding-mode disturbance observer(ASMDO)-based controller for enhanced robustness and tracking accuracy.
Overall, to ensure accurate and robust operation of AMS, it is crucial to compensate for the large aggregate perturbations caused by the manipulator, which is typically achieved through observers \cite{emami2021simultaneous,chen2022adaptive,cao2023eso}, feed-forwards \cite{li2022adaptive,8924899}, and methods adept at handling strong nonlinearities, such as neural networks \cite{cao2022adaptive}.

The coupling disturbance caused by pose changes in manipulators are the primary source of aggregated disturbances in AMS,as pointed out by Li et al. \cite{li2022adaptive}.This is mainly attributed to three factors: shifts in the system's COM, coupling disturbance forces, and coupling disturbance torques.Manipulators with motors and reducers in their joints, intensify this issue by adding substantial weight, thereby amplifying inertial forces and torques during motion~\cite{mortl2012role}. Even with desktop-level manipulator platforms, balancing load capacity with coupling disturbance remains a challenge due to AMS's payload constraints. 

This paper considers adopting a structural design method to alleviate coupling disturbances in AMS. The essence of this approach involves physically placing the actuators at the proximal end of the manipulator and driving the distal joints through a transmission medium. This design aims to minimize inertia and centralize the overall mass distribution, effectively reducing the shift in the system's center of gravity and the coupling disturbances during the arm's movement. It is expected to enhance the dynamic motion performance of the system.

Contrary to traditional serial manipulators that position heavy actuators at the joints, cable-driven serial manipulators((alternatively referred to as tendon-driven)) utilize flexible cables and centralize the mechatronic servo system at the base \cite{chen2023kinematics,sefati2020surgical, kim2017anthropomorphi}. This design not only reduces joint weight and dimensions but also enhances waterproofing and dustproofing \cite{ozawa2013analysis}. Such a configuration is advantageous for creating lightweight and compact end effectors, especially when intricate multi-DoF motions are required in confined spaces \cite{li2020modularization,dou2022inverse,chien2023design}. A notable example of this approach is the Da Vinci surgical robot \cite{gu2023survey,kaouk2017atlas,hong2013design}.


Cable-pulley systems are prevalent in the field of cable-driven serial manipulators\cite{wang2020adaptive}. Force amplification with tendons is straightforward due to the adjustable speed reduction ratio. These tendons, often made of stainless steel or polymer cables, transfer forces from motors to joints, enhancing the safety of robot-human interactions \cite{li2021development, li2020modularization}. Their inherent compliance and dynamics make them suitable for mimicking musculoskeletal characteristics, leading to their popularity in anthropomorphic robots.However, a notable challenge with the single winding of steel cables is their reduced stiffness and torque output, especially when compared to traditional industrial manipulators. To ameliorate this deficiency,Kim introduced a tension-amplification mechanism \cite{kim2017anthropomorphi,kim2018quaternion}  for a 7-DOF manipulator. This innovation not only achieves industrial-level joint stiffness and torque output but also ensures safer human-robot interactions due to the reduced mass and inertia of the designed robotic arm.

Based on the analysis above, equipping the AMS with a cable-driven serial manipulator, integrated with a tension amplification mechanism, strikes an optimal balance between load capacity and minimization of coupling disturbances. The cable-driven approach in AMS brings distinct advantages, primarily attributed to its inherent compliance and reduced inertia, a result of distributed actuator placement. This configuration effectively minimizes coupling disturbances. Furthermore, numerical optimization tools such as topology optimization and internal lattice structures can be employed to significantly reduce structural weight of the manipulator. With the advancements in additive manufacturing technologies, it is now feasible to realize complex lightweight structures derived from topology and lattice optimizations \cite{NEJAT2021106727,LI2022107457}.

\begin{figure}[htbp]
  \centering
  \includegraphics[width=0.5\textwidth]{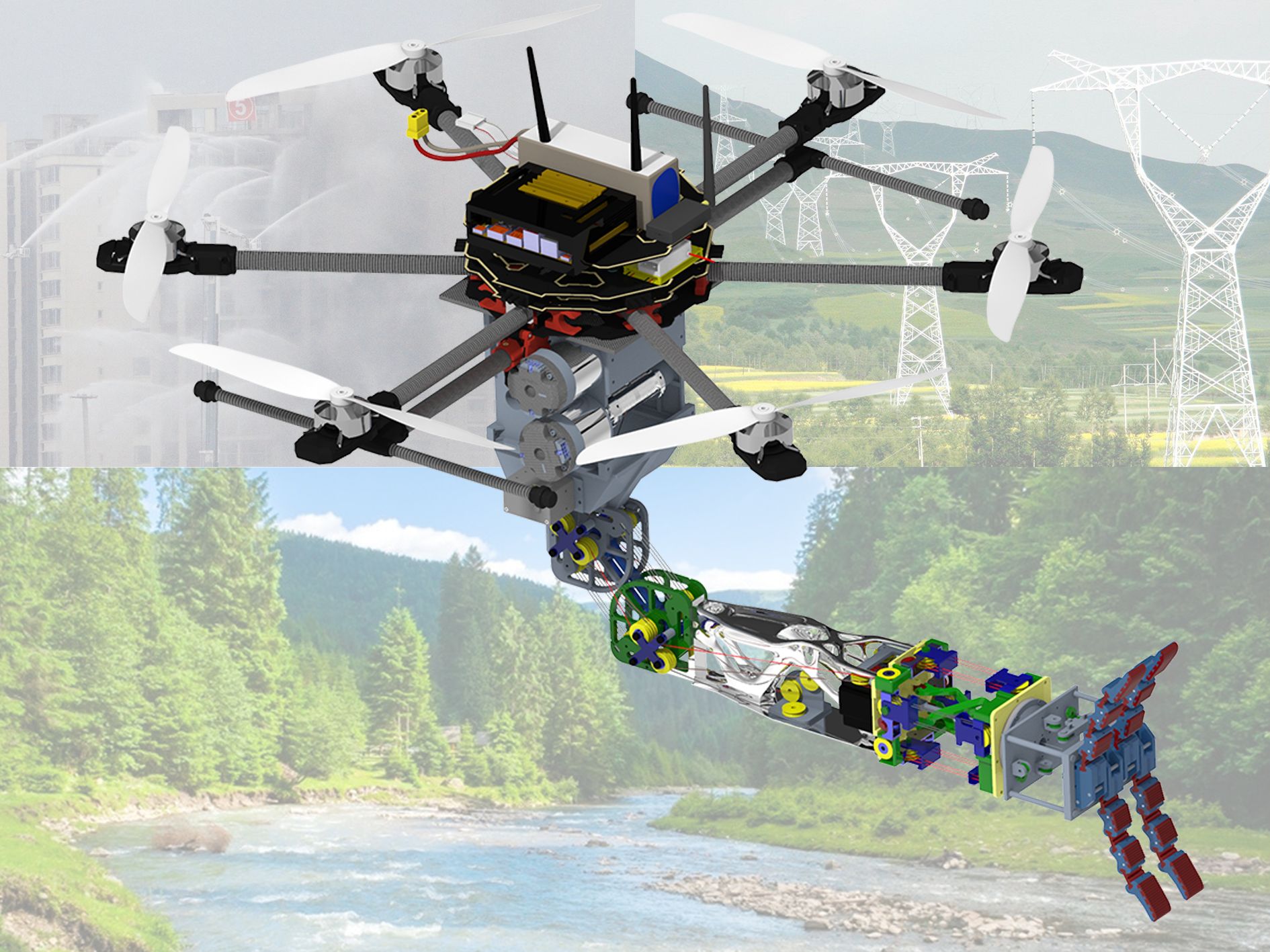}
  \caption{The proposed aerial tendon-driven manipulator(ATDM)}
  \label{figure:The proposed aerial tendon-driven manipulator(ATDM)}
    \small\textit{Note:JPG format, 1-column fitting image}
\end{figure}

This paper presents the novel ATDM, illustrated in Fig.(\ref{figure:The proposed aerial tendon-driven manipulator(ATDM)}). This design combines a hexrotor with a 4-DOF tendon-driven manipulator. By positioning all actuators at the base of the arm, we achieve a dual benefit: a significant reduction in coupling disturbance and a more efficient integration of mechatronics, particularly beneficial in adverse environments. Leveraging topology optimization and finite element methods, we optimized the design of the moving joints and links, striking a balance between stiffness and weight reduction. The system's total weight is approximately 2.7 kg, with the dynamic portion of the arm contributing only 0.818 kg, exemplifying an optimized weight distribution.

The primary contributions of this paper are outlined as follows:
\begin{enumerate}
  \item The anthropomorphic tendon-driven manipulator was first applied in AMS, with all electromechanical servo systems were located at the base,which effectively mitigating the issue of excessive coupling disturbance within AMS. The design process of the manipulator is guided by the dynamics of AMS and an accurate model of coupling disturbance. Semi-physical simulation has demonstrated that the tendon-driven approach presents a notable reduction in disturbance, outperforming traditional serial manipulator setups.
  
  \item The detailed manipulator structural design and its optimization process are demonstrated. Including tension-amplification tendon(TAT) that can increase joint stiffness and thus increase load capacity; 1-DOF and 2-DOF joints based on TAT; lightweight tensioning mechanisms with adjustable tension; and finite element topology optimization, lattice optimization in the process of weight reduction of the manipulator.
  
  \item A kinematic model of the arm is formulated, supplemented by workspace analysis of the ATDM system. In addition, the output torque and stiffness of each joint of the arm based on the TAT mechanism are also derived.
  
\end{enumerate}

The remainder of this paper is structured as follows: Section \ref{section:2} deduces the coupling disturbance model of AMS.Section \ref{section:3} shows the mechanical design and provides an analysis. Section \ref{section:4} details the kinematics, analyzes the output torque and stiffness of joints. Section \ref{section:5} presents the semi-physical simulation results. Finally, Section \ref{section:6} offers conclusions drawn from the research.

\section{Modeling}\label{section:2}

This section presents the mathematical model of AMS, derives the coupling disturbance model and compare the impact of various actuator distributions,and further outlines the design objectives of the ATDM.

\subsection{Dynamics}\label{subsection:2.1}

The AMS studied in this article includes a hexrotor UAV and a multi-DOF manipulator. The manipulator is composed of multiple rigid links connected by joints. The end effector is mounted on the last link of the manipulator (refer to Fig.(\ref{figure:The proposed aerial tendon-driven manipulator(ATDM)})).

\begin{figure}[htbp]
  \centering
  \includegraphics[width=0.5\textwidth]{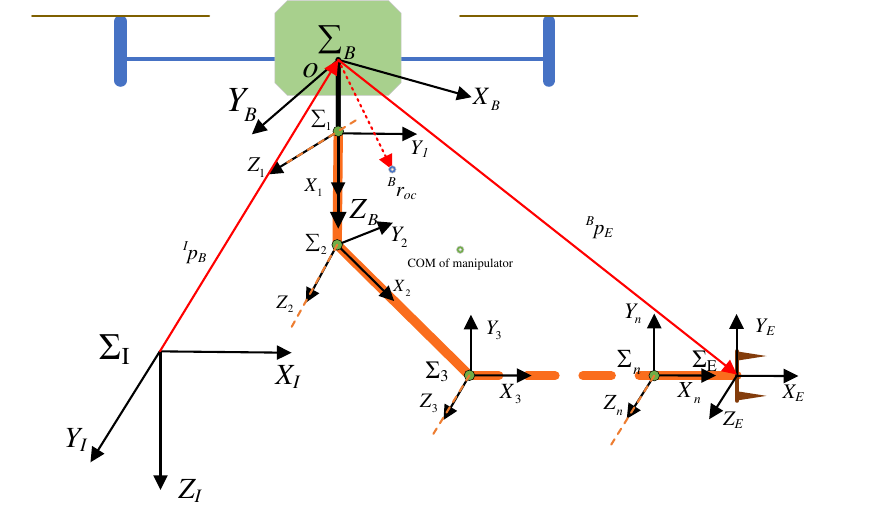}
  \caption{The coordinate frame of AMS}
  \label{figure:coordinate_frame}
  \small\textit{Note:PDF format, 1-column fitting image}
\end{figure}

The AMS coordinate system has been established and demonstrated in Fig.\ref{figure:coordinate_frame},where the North, East, Down (NED) coordinate system ${\Sigma _I}$ and ${\Sigma _B}$ represent the inertial coordinate system and the body coordinate system built at the center of mass of the UAV, respectively. ${}^I\boldsymbol{\mathit{p}}_B \in \boldsymbol{\mathit{R}}^3$ denote the COM of UAV  in ${\Sigma _I}$, and let ${}^B\boldsymbol{\mathit{p}}_E \in \boldsymbol{\mathit{R}}^3$ represent the end-effector in ${\Sigma _B}$. The position, velocity, and angular velocity of the end-effector in the ${\Sigma _I}$ can be derived as
\begin{equation}
    \begin{array}{l}
    {}^I\bm{p}_E = {}^I\bm{p}_B + {}^I\bm{R}_B{}^B\bm{p}_E\\
    {}^I\bm{\dot{p}}_E = {}^I\bm{\dot{p}}_B - \text{Skew}\left( {}^I\bm{R}_B{}^B\bm{p}_E \right){}^I\bm{\omega}_B + {}^I\bm{R}_B{}^B\bm{\dot{p}}_E\\
    {}^I\bm{\omega}_E = {}^I\bm{\omega}_B + {}^I\bm{R}_B{}_E^B\bm{J}_{wq}\bm{\dot{q}}_m\\
    {}^B\bm{\dot{p}}_E = {}_E^B\bm{J}_{vq}\bm{\dot{q}}_m
\end{array}
\end{equation}
where $ \text{Skew}( \cdot )$ is the anti-symmetric matrix, ${\dot {\boldsymbol{\mathit{q}}}_m}$ represents the joint angular velocities of the manipulator, ${}_E^B{\boldsymbol{\mathit{J}}_{wq}}$  maps the joint angular velocities of the manipulator to the angular velocity of ${\Sigma _E}$ with respect to ${\Sigma _B}$, ${}_E^B{\boldsymbol{\mathit{J}}_{vq}}$ is the Jacobian matrix relating the linear velocity of ${\Sigma _E}$ with respect to ${\Sigma _B}$ to the manipulator's joint angular velocities, and ${}^I\boldsymbol{\mathit{R}}_B \in SO(3)$  is the representation of attitude transformation matrix of ${\Sigma _B}$ in the ${\Sigma _I}$. When $\boldsymbol{\mathit{\Phi}}  = {[\phi ,\theta ,\psi ]^T}\in \boldsymbol{\mathit{R}}^3$ correspond to the UAV's pitch, roll, and yaw angles, respectively,it can be derived as
{
\begin{equation}
{}^I{\bm{R}_B} = \left[ {\begin{array}{*{20}{c}}
{C\theta C\psi }&{S\phi S\theta C\psi  - C\phi S\psi }&{C\phi S\theta C\psi  + S\phi S\psi }\\
{C\theta S\psi }&{S\phi S\theta S\psi  + C\phi C\psi }&{C\phi S\theta S\psi  - S\phi C\psi }\\
{ - S\theta }&{S\phi C\theta }&{C\phi C\theta }
\end{array}} \right]
\end{equation}
}
where, $C(\cdot)$ denotes $\cos( \cdot )$, and $S(\cdot)$ denotes $\sin(\cdot)$. 

Based on Newton's Second Law and Euler's equations of motion, the dynamic equation of the AMS can be formulated as
\begin{align}
    {}^I\bm{\dot{P}}_B =& {}^I\bm{v}_B, \notag\\
    {}^I\bm{\dot{v}}_B =& \bm{g} - \frac{{{}^I\bm{f} - {}^I\bm{f}_{dis}}}{{m_{UAV} + m_{MAN}}}, \notag\\
    {}^I\bm{\dot{R}}_B =& {}^I\bm{R}_B \cdot \text{Skew}({}^B\bm{\omega}_b) = \text{Skew}({}^I\bm{\omega}_b) \cdot {}^I\bm{R}_B, \notag\\
    {}^B\bm{\dot{\omega}}_b =& \frac{ {}^B\bm{\tau} - {}^B\bm{\omega}_b \times (({}^B\bm{I}_{UAV} + {}^B\bm{I}_{MAN}){}^B\bm{\omega}_b) + {}^B\bm{\tau}_{dis} }{({}^B\bm{I}_{UAV} + {}^B\bm{I}_{MAN})} 
\end{align}
where, ${}^I{\boldsymbol{\mathit{v}}_B}\in {\boldsymbol{\mathit{R}}}^3$ represents the velocity vector of the AMS in ${\Sigma _I}$, $\boldsymbol{\mathit{g}}$ is the gravity vector, ${}^I\boldsymbol{\mathit{f}}={}^I\boldsymbol{\mathit{R}}_B (0,0,f)^T$ signifies the lift generated by the UAV's propellers, $m_{UAV}$ and $m_{MAN}$ represent the masses of the UAV and the manipulator respectively, ${}^B{\boldsymbol{\mathit{I}}_{UAV}}$ and ${}^B{\boldsymbol{\mathit{I}}_{MAN}}$ are the inertia tensor matrices of the UAV and the manipulator arm in ${\Sigma _B}$ respectively, ${}^B{\boldsymbol{\mathit{\omega}} _b}$ denotes the UAV's angular velocity in ${\Sigma _B}$ , ${}^B\boldsymbol{\mathit{\tau}}$  describes the expression of the torque generated by the UAV's propellers in ${\Sigma _B}$, ${}^I{\boldsymbol{\mathit{f}}_{dis}}\in {\boldsymbol{\mathit{R}}}^3$ and ${}^B{\boldsymbol{\mathit{\tau}} _{dis}}\in {\boldsymbol{\mathit{R}}}^3$ represent the coupling disturbance force and torque exerted on the UAV by the motion of the manipulator.
\subsection{Coupling Disturbance}\label{subsection:2.2}

The coupling disturbance model has been precisely modeled by Li et al. \cite{li2022adaptive}. Considering that understanding the aforementioned variables is crucial for a complete representation of the system's dynamics, this paper extends that model to analyze the effect of actuator mass distribution on the coupling disturbance. The details are as follows:
\begin{align}
    \bm{F}_{\text{dis}} &=  - (m_{UAV} + m_{MAN}) {}^I \bm{R}_B \Bigl( {}^B \bm{\omega}_b \times \left( {}^B \bm{\omega}_b \times {}^B \bm{r}_{oc} \right) \notag\\
    &\quad+ {}^B \bm{\dot{\omega}}_b \times {}^B \bm{r}_{oc} + 2 {}^B \bm{\omega}_b \times {}^B \bm{\dot{r}}_{oc} \Bigl), \notag\\
    {}^B \bm{\tau}_{dis} &=(m_{UAV} + m_{MAN}) \Bigl( {}^B \bm{r}_{oc} \times {}^B \bm{R}_I (\bm{g} \bm{e}_3 - \bm{\dot{v}}_b)  + {}^B \bm{\ddot{r}}_{oc} \times {}^I \bm{R}_B {}^I \bm{\dot{p}}_B \Bigl) \notag\\
    &\quad- {}^B \bm{I}_{UAV} {}^B \bm{\dot{\omega}}_b - {}^B \bm{\omega}_b \times \left( {}^B \bm{I}_{MAN} {}^B \bm{\omega}_b \right)  - {}^B \bm{\dot{I}}_{MAN} {}^B \bm{\omega}_b \notag\\
    &\quad- \frac{(m_{UAV} + m_{MAN})^2}{m_{MAN}} \Bigl( {}^B \bm{r}_{oc} \times {}^B \bm{\ddot{r}}_{oc}  + {}^B \bm{\omega}_b \times \left( {}^B \bm{r}_{oc} \times {}^B \bm{\dot{r}}_{oc} \right) \notag\\
    &\quad+ {}^I \bm{R}_B {}^I \bm{\dot{p}}_B \times {}^B \bm{\dot{r}}_{oc} + {}^I \bm{R}_B {}^I \bm{\dot{p}}_B \times ({}^B \bm{\omega}_b \times {}^B \bm{\dot{r}}_{oc}) \Bigl)
\end{align}
where
\begin{subequations}
    \begin{equation}  
    {}^B\bm{r}_{oc} = \frac{1}{{m_{UAV} + m_{MAN}}}\sum\limits_{i = 1}^n {(m_{Li}{}^B\bm{p}_{cLi}  + m_{Mi}{}^B\bm{p}_{cMi})}
    \label{equation:coupling disturbance_roc}
    \end{equation}
    \begin{align}
    {}^B\bm{I}_{MAN} &= \sum_{i = 1}^n \bigg( {}^B\bm{R}_{Li} \bm{I}_{Li}^{\Sigma_{Li}} {}^B\bm{R}_{Li}^{-1} + {}^B\bm{R}_{Mi} \bm{I}_{Mi}^{\Sigma_{Mi}} {}^B\bm{R}_{Mi}^{-1} \notag\\
    &\quad+ m_{Li} \left( \left\| {}^B\bm{p}_{cLi} \right\|^2 \bm{I}_{3 \times 3} - {}^B\bm{p}_{cLi} {}^B\bm{p}_{cLi}^T \right)  \notag\\
    &\quad+ m_{Mi} \left( \left\| {}^B\bm{p}_{cMi} \right\|^2 \bm{I}_{3 \times 3} - {}^B\bm{p}_{cMi} {}^B\bm{p}_{cMi}^T \right) \bigg)
    \label{equation:coupling disturbance_iman}
    \end{align}
\end{subequations}

The symbol ${}^B{\boldsymbol{\mathit{r}}_{oc}}\in {\boldsymbol{\mathit{R}}}^3$ represents the COM of the AMS in the ${\Sigma _B}$ coordinate system. $m_{Li}$,$m_{Mi}$,${}^B\boldsymbol{\mathit{p}}_{cLi}\in {\boldsymbol{\mathit{R}}}^3$, and ${}^B\boldsymbol{\mathit{p}}_{cMi}\in {\boldsymbol{\mathit{R}}}^3$ represent the mass and the respective center-of-mass locations of the $i{\text{-th}}$ link structure (excluding the actuator) and the $i{\text{-th}}$ joint actuator in the  ${\Sigma _B}$ coordinate system. ${}^B{\dot {\boldsymbol{\mathit{r}}}_{oc}}$ and ${}^B{\ddot {\boldsymbol{\mathit{r}}}_{oc}}$ are its first and second derivatives, respectively. In the expression of ${}^B{\boldsymbol{\mathit{I}}_{MAN}}$, ${}^B{\boldsymbol{\mathit{R}}_{Li}}$, ${}^B{\boldsymbol{\mathit{R}}_{Mi}}$, $^B{\boldsymbol{\mathit{p}}_{cLi}}$, and $^B{\boldsymbol{\mathit{p}}_{cMi}}$ represent the configurations and coordinate system origins of the $i{\text{-th}}$ link structure (excluding the actuator) and the $i{\text{-th}}$ joint actuator in ${\Sigma _B}$, established at their respective centers of mass. $\boldsymbol{\mathit{I}}_{Li}^{{\Sigma _{Li}}}$ and $\boldsymbol{\mathit{I}}_{Mi}^{{\Sigma _{Mi}}}$ represent the inertia tensor matrices of the $i{\text{-th}}$ link structure and the $i{\text{-th}}$ joint actuator in their respective center-of-mass coordinate systems.  ${}^B{\dot{\boldsymbol{\mathit{ I}}}_{MAN}}$ is the derivative of ${}^B{\boldsymbol{\mathit{I}}_{MAN}}$.

The aforementioned elements model each link and actuator in the manipulator separately, allowing for better calculations of the impact of actuator weight distribution on coupling disturbance.

Movement of the manipulator leads to changes  ${}^B{\boldsymbol{\mathit{r}}_{oc}}$ and ${}^B{\boldsymbol{\mathit{I}}_{MAN}}$ of the AMS, along with associated variables. The faster the movement of the arm, the larger these variables become, thereby increasing the resultant coupling disturbance.

To  places the actuators of the manipulator on a base that is fixed to the UAV. In this case, terms related to the actuators in the derivatives of ${}^B{\boldsymbol{\mathit{r}}_{oc}}$ and ${}^B{\boldsymbol{\mathit{I}}_{MAN}}$ become zero, significantly reducing the coupling disturbance when the manipulator is in motion. Conversely, traditional manipulator mounted on flying platforms have heavier actuators that are typically distributed around the joints. This makes $m_{Mi}$, ${}^B\boldsymbol{\mathit{p}}_{cMi}$, and ${}^B{\boldsymbol{\mathit{I}}_{MAN}}$ not only nonzero but also dominant terms in the Equ.(\ref{equation:coupling disturbance_roc})and (\ref{equation:coupling disturbance_iman}), significantly increasing their coupling disturbance. 

\subsection{Design Objectives}

The ATDM(refer to Fig.(\ref{figure:The proposed aerial tendon-driven manipulator(ATDM)})) designed in this paper aims to address several challenges in existing manipulators of AMS, specifically when it comes to tasks involving aerial manipulation. We focus on two primary issues commonly encountered in existing AMS:

\begin{itemize}
    \item \textbf{Coupling Disturbance:} Traditional AMS with serial manipulators often encounter large unwanted coupling disturbance to the UAV, due to the placement of actuators and the overall COM. As an under-actuated platform, the lift provided by each blade of the UAV is relatively limited, and the control authority is easily saturated.

    This problem is mitigated by introducing a novel tendon-driven manipulator, which remotely drives joint rotation via actuators mounted at the UAV's base. This design significantly reduces coupling disturbance and is further optimized through finite element analysis to achieve a lightweight construction.

    \item \textbf{Payload Capacity:} The payload capacity of AMS equipped with conventional manipulator systems is limited due to the load capacity of multi-rotor UAVs, resulting in restricted joint stiffness and payload capabilities. These limitations make it difficult for the arms to be applied in scenarios requiring the handling of large loads. While industrial manipulators address payload issues, they compromise energy efficiency due to their higher weight-to-payload ratio.

    We address this issue by employing a TAT mechanism to increase the overall stiffness and output of the manipulator.
\end{itemize}

\section{Mechanical Design of the Manipulator}\label{section:3}

The COM of the serial manipulator often equiped by AMS is often far away from the COM of the drone. This setup will result in increased coupling disturbance forces and torques during arm movement. ATDM positions the electromechanical servo system of the manipulator on the base, effectively reducing the coupling disturbance during movement. The structure was further optimized through finite element analysis to achieve a lightweight, low-inertia manipulator that is both compatible and strong, thereby enhancing the operational capabilities of the AMS.

This section presents the structural design of the ATDM system's manipulator, accompanied by the configuration of the tendons, the motion analysis,and the distribution of actuators. In addition, finite element analysis was conducted to achieve weight reduction.

\subsection{Tension amplification tendon mechanism}\label{subsection3.1}

Traditional "cable-pulley" based cable-driven manipulators, exemplified by the Da Vinci surgical robots \cite{gu2023survey,kaouk2017atlas,hong2013design}, utilize a single steel cable as the tendon. While these designs offer benefits like vibration damping and lightweight construction, they suffer from low stiffness and limited torque output, making them unsuitable for heavy loads.  Sava \footnote{Sava. URL: \href{https://www.savacable.com}{https://www.savacable.com}. Accessed on: [2023/10/31].} suggests that the winding diameter of the steel cable (i.e., the radius of the capstan or pulley) should be more than 18 times the cable's outer diameter.Our practical tests revealed that using thicker steel cables, coupled with larger pulleys, significantly increases the manipulator's volume and the pulley's weight, although the system gains greater stiffness and breaking strength. For cable-driven manipulators that employ numerous pulleys, a trade-off between these factors is evidently needed.

\begin{figure}[htbp]
  \centering
  \includegraphics[width=0.5\textwidth]{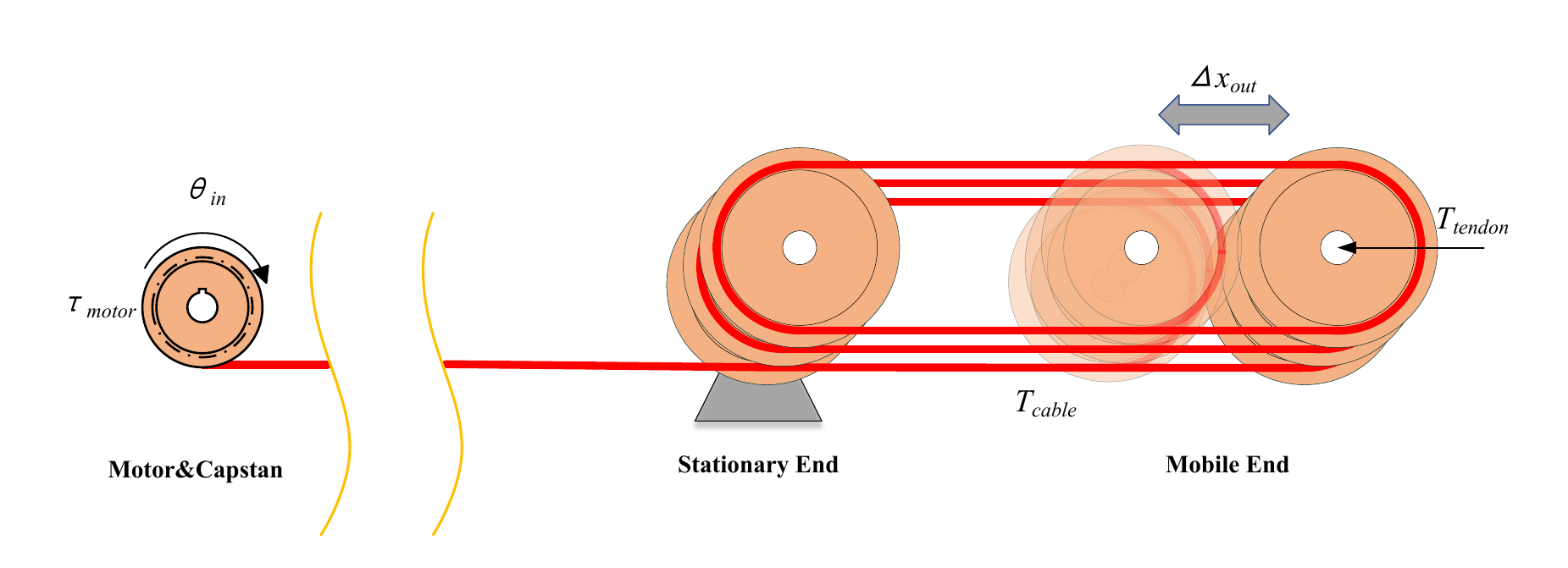}
  \caption{Tension-amplification tendon}
  \label{figure:Tension-amplification tendon}
  \small\textit{Note:PDF format, 1-column fitting image}
\end{figure}

ATDM leverages a TAT mechanism at each joint of the manipulator, as shown in Fig.\ref{figure:Tension-amplification tendon}. ATDM uses steel cables with diameter of 0.6mm, with a $7\cdot19$ structure and an external nylon coating, at every joint. Because of the small cross-sectional area of the selected cable and its high flexibility, the cable is guided through multiple grooved pulleys at both the driving and passive ends of each joint of the arm, winding repeatedly,this makes up the TAT structure. TAT structure increases the arm's stiffness and load-bearing capability. The flexible cables, winding back and forth, form a TAT system that acts as a reducer at the joints, enhancing joint stiffness, improving the precision and dynamic performance of the manipulator.

Based on the TAT mechanism, when the motor at the fixed end inputs a torque of $\tau_{motor}$, the total output pulling force of the tendon is 
\begin{equation}
    T_{tendon} = N \cdot \tau_{motor} / R_{capstan} \label{equation1}
\end{equation}
where,$R_{capstan}$ is the radius of the cablecapstan, and N is the number of times the cableis wound around the TAT mechanism, which is twice the number of pulley groups at the moving end.

Based on the virtual works principle, the reduction ratio is $N$, and the tendon stiffness $E_{tendon}$ can be derived as
\begin{align}
    E_{tendon} &= T_{tendon} / \delta x_{tendon} \notag\\
    &= N \cdot T_{cable} / (\delta x_{cable}/N) \notag\\
    &= N^2 E_{cable} \label{equation2}   
\end{align}

where $\delta x_{tendon}$ is the deformation of the tendon when the output force is $T_{tendon}$, $T_{cable}$ is the tension of the steel cable at this time, and $x_{cable}$ is the deformation of the entire cable. Based on Eq.(\ref{equation2}) , the stiffness of the tendon is $N^{2} $ times that of a single cable.

\subsection{Design of Anthropomorphic Elbow Joint}\label{section:3.2}

\subsubsection{Mechanical Design}
Drawing inspiration from human anatomy \cite{agur2009grant}, the human elbow is a 1-DOF rotational joint primarily controlled by the antagonistic action of the biceps brachii and triceps brachii. The contraction of one muscle group results in the elongation of the other, facilitating precise elbow movement.

\begin{figure}[h]
  \centering
  \includegraphics[width=0.5\textwidth]{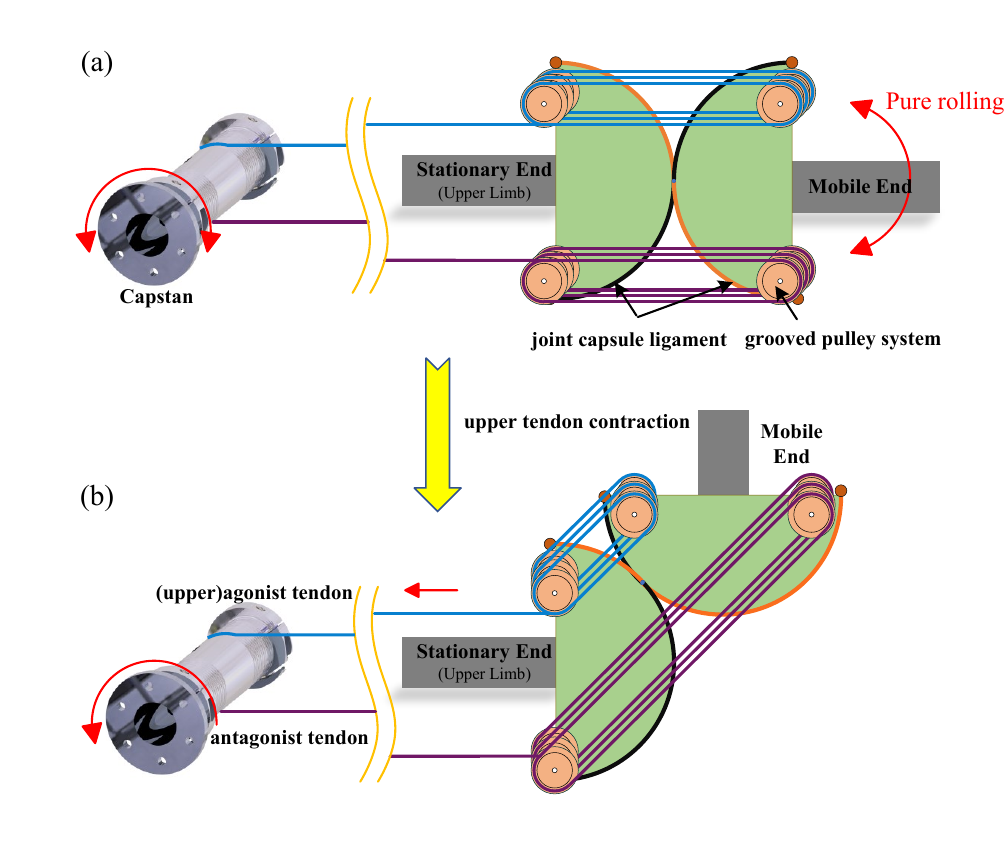}
  \caption{Elbow Joint Utilizing TAT.(a)Initial state.(b)Bending state}
  \label{figure:Elbow Joint Utilizing Tension-Amplification Mechanism}
  \small\textit{Note:PDF format, 1-column fitting image}
\end{figure}

Building on the TAT concept from section \ref{subsection3.1}, we emulate this antagonistic mechanism for the robotic elbow joint using a pair of opposing TATs. As illustrated in Fig.(\ref{figure:Elbow Joint Utilizing Tension-Amplification Mechanism}), when the joint rotates counterclockwise, the upper tendon acts as the agonist, with the lower tendon as the antagonist, and vice versa for clockwise movement. For smooth, dead-zone-free motion, these tendons must synchronize in real-time and remain taut. While separate actuators could drive them, it might add to the manipulator's weight and inertia.

To address this issue, we adopted an innovative cable drive scheme, as sourced from the literature \cite{kim2017anthropomorphi}. In our mechanical design, particular attention was given to the mechanism of motion translation between the forearm and upper arm. Specifically, we engineered a unique set of circular contact surfaces that directly translate the movements of the forearm and upper arm into pure rolling motion between these circular surfaces.

The key to this design lies in its specialized geometric structure. We meticulously designed the radii and curvatures of the contact surfaces to ensure that as the forearm and upper arm rotate or bend, the interaction between these surfaces results in a slip-free pure rolling motion. The advantage of this pure rolling mechanism is that it significantly reduces energy loss due to friction and sliding, thereby enhancing the mechanical efficiency of the entire system.

Furthermore, this design also takes into account the length variation of the agonist and antagonist tendons. In traditional tendon-driven systems, tendons undergo elongation and shortening during movement, leading to energy loss and reduced efficiency. However, in our design, as the motion of the forearm and upper arm is translated into pure rolling between circular surfaces, this ensures that the total length of the agonist and antagonist tendons (the tendons driving and controlling limb movement) remains constant throughout the range of motion. This constancy in length not only reduces tendon wear but also enhances the system's response speed and energy transmission efficiency.

To achieve a pure rolling motion in the elbow joint, a mechanism akin to the joint capsule is essential for stability and to prevent slippage. While various mechanisms like antiparallelograms and gear meshing-based transmissions exist, this study opts for cables, mimicking ligaments' role in the joint capsule, especially considering its aerial application and the need for impact resistance.

Our design employs two intersecting steel cables to ensure pure rolling, prevent slippage, and bolster the elbow joint's torsional rigidity. The fixed pulleys from the antagonistic TAT structures are symmetrically placed relative to the rolling surface's curvature center of the proximal link. Similarly, the moving pulleys are symmetrically positioned concerning the distal link's rolling surface. This symmetrical arrangement facilitates uniform tendon motion, allowing both cable sections to wind on a single capstan driven by one actuator, as shown in Fig. (\ref{figure:Elbow Joint Utilizing Tension-Amplification Mechanism}). Such a design effectively reduces the AMS's coupling disturbance.

\subsubsection{Kinematic Analysis of Elbow Joint}\label{section:3.2.2}
This section analyzes the movement of the elbow joint. Without loss of generality, we consider the situation where the tendon on the top acts as the agonist .

\begin{figure}[htbp]
  \centering
  \includegraphics[width=0.5\textwidth]{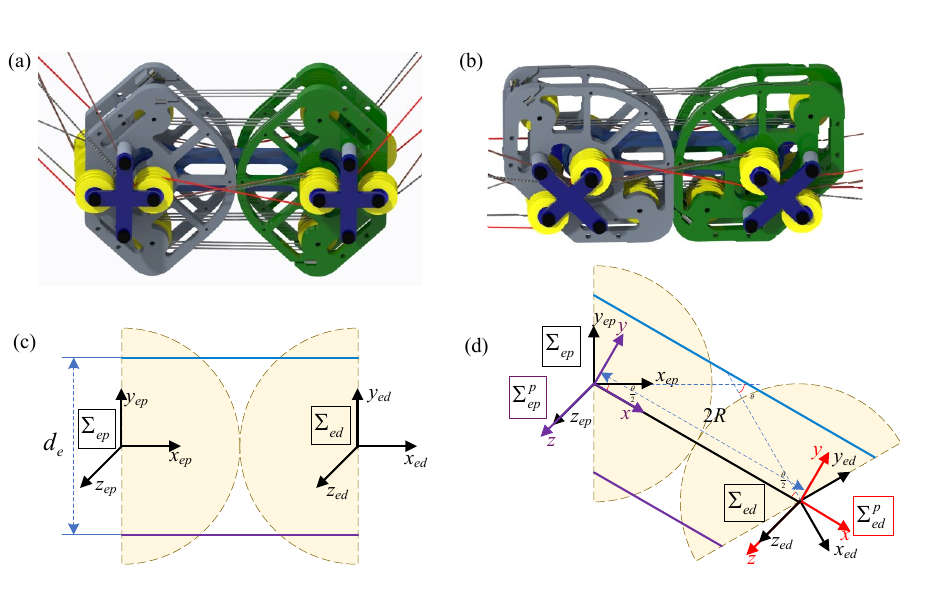}
  \caption{Elbow joint design detail and analysis. (a)Detail(initial state).\\(b)Detail(Bending state).(c)Schematic diagram(initial state).(d)Schematic diagram(Bending state)}
  \label{figure:Elbow joint design and analysis}
  \small\textit{Note:PDF format, 1-column fitting image}
\end{figure}

The design details and simplified model of the elbow joint is shown in Fig.(\ref{figure:Elbow joint design and analysis}). As it is a pure rolling joint, it can be decomposed into two motion-coupled serial rotational pairs for analysis.

Coordinate systems are established at the center of the fixed end and the center of the moving end of the elbow joint, respectively, labeled as ${\Sigma _{ep}}\{ {O_{ep}} - {X_{ep}}{Y_{ep}}{Z_{ep}}\}$ and ${\Sigma _{ed}}\{ {O_{ed}} - {X_{ed}}{Y_{ed}}{Z_{ed}}\} $. Due to the coupling effects inherent in the joint mechanism, intermediate coordinate systems $\Sigma _{ep}^p$ and $\Sigma _{ed}^p$ are introduced to facilitate the description of the homogeneous transformations between ${\Sigma _{ep}}$ and ${\Sigma _{ed}}$,where the relationship of transformations is ${\sum _{ep}} \to \sum _{ep}^p \to \sum _{ed}^p \to {\sum _{ep}}$. For easier identification and readability, colored bounding boxes have been added around the frames of these intermediate coordinate systems in Fig.\ref{figure:Elbow joint design and analysis}(d). The colors correspond to the axes of their respective coordinate systems.

Considering that the agonist contracts (colored blue, label antagonistic as purple in parallel), and the angle of the elbow joint rotation is $\theta$, with a curvature radius of $R$ and the distance between the agonist and antagonist muscles as $d_e$. Refer to geometric relationships in in Fig.\ref{figure:Elbow joint design and analysis}(c) and Fig.\ref{figure:Elbow joint design and analysis}(d) can derive the lengths of the active tendon and the antagonist tendon at this time as
\begin{equation}
    L_{ago} = N_e(2R - d_e\sin (\theta/2)); L_{ant} = N_e(2R + d_e\sin (\theta/2)) \label{equation:3}
\end{equation}

where $N_e$ is the number of windings of the tendon of the elbow joint tension magnifier, the changes in length of the active tendon and the antagonist tendon are respectively
\begin{equation}
    \Delta L_{ago} = - N_e d_e \sin (\theta/2); \Delta L_{ant} = N_e d_e \sin (\theta/2)
\end{equation}

It can be observed that as the angle $\theta$ changes, the changes in length of the active tendon and the antagonist tendon can just compensate for each other. Meanwhile, the homogeneous transformation relationship from the fixed end center of the elbow joint to the moving end of the elbow joint is
\begin{equation}
    \begin{array}{l}
^{ep}\bm{T}_{ed} = Rot(Z, \theta/2) Trans(X, 2R) Rot(Z, \theta/2)\\
\quad = \left[ \begin{array}{cccc}
C(\theta) & -S(\theta) & 0 & 2R \cdot C(\theta/2) \\
S(\theta) & C(\theta) & 0 & 2R \cdot S(\theta/2) \\
0 & 0 & 1 & 0 \\
0 & 0 & 0 & 1
\end{array} \right]
\end{array}
\end{equation}

\subsection{Design of Anthropomorphic Wrist Joint}\label{Section:3.3}

\subsubsection{Wrist Joint Design Analysis}

According to systematic anatomy \cite{agur2009grant}, the human wrist joint movements can be categorized as follows: 
\begin{itemize}
  \item Forearm pronation and supination are mainly led by the \textit{pronator teres} and \textit{supinator muscles}, respectively.
  \item Wrist flexion and extension are principally driven by the \textit{flexor carpi radialis} for flexion and \textit{extensor carpi radialis} for extension.
  \item Radial and ulnar deviations are primarily propelled by the \textit{flexor carpi radialis} for radial deviation and \textit{flexor carpi ulnaris} for ulnar deviation.
\end{itemize}

The anthropomorphic manipulator wrist joint design comprises a 2-DOF parallel joint for wrist flexion-extension and ulnar-radial deviation, with a serial joint for forearm pronation and supination. Adopting a design approach akin to the elbow joint, six tendons drive the wrist motion. Through kinematics design, each antagonistic tendon pair maintains consistent length changes, allowing for three actuator-driven tendon sets. The parallel joint's design draws inspiration from the quaternion joint \cite{kim2018quaternion}, a 2-DOF parallel mechanism using an antiparallelogram linkage, achieving a pure rolling motion on a spherical surface.

\subsubsection{Wrist Joint Design}

A single parallelogram mechanism can only approximate an arc of a circle in its plane(refer to \ref{appendix:anti}). By positioning two sets of parallelogram mechanisms perpendicularly intersecting at their centers, and substituting the rotation pairs at the connections of each direct link with the moving and static platforms with Hooke's joints, the intersection points of the parallelogram linkages lies on an ellipsoidal surface during movement as
\begin{equation}
    \frac{x^2}{(l/2)^2} + \frac{(y-h)^2}{(l/2)^2 - w^2} + \frac{z^2}{(l/2)^2} = 1
\end{equation}

The assembled mechanism contains redundant parallel constraints. By removing one of the support chains and positioning the remaining three at equal angular intervals, it can be simplified. Analyzing its degrees of freedom through screw theory reveals that it has 2 degrees of freedom in the initial configuration and none in the bent state. Although the moving platform's constraint wrench system is full rank, it is approximately linearly dependent. Theoretically, this configuration is fully constrained; however, due to minor deformations of the linkage under slight stress, it can tilt in two directions\cite{shikata2014}. Analysis of the mechanism's movement using the atlas method suggests that, although the moving platform constraint force line has six, three of them nearly intersect \cite{kim2018quaternion}. In the presence of minor errors, a phenomenon of dimensionality reduction occurs. To verify its movement capabilities, this study performed an explicit dynamics analysis in \textit{Abaqus} (refer to \ref{appendix:quaternion}).

Minor errors during machining and slight deformation of the mechanism under bending forces reduce the dimension of this mechanism's constraint wrench system and constraint force line. The quaternion joint can deflect and is a two-degree-of-freedom spherical pure rolling joint. It has a smooth working space, with no singular points within the workspace and continuous boundaries.

\begin{figure}[htbp]
  \centering
  \includegraphics[width=0.5\textwidth]{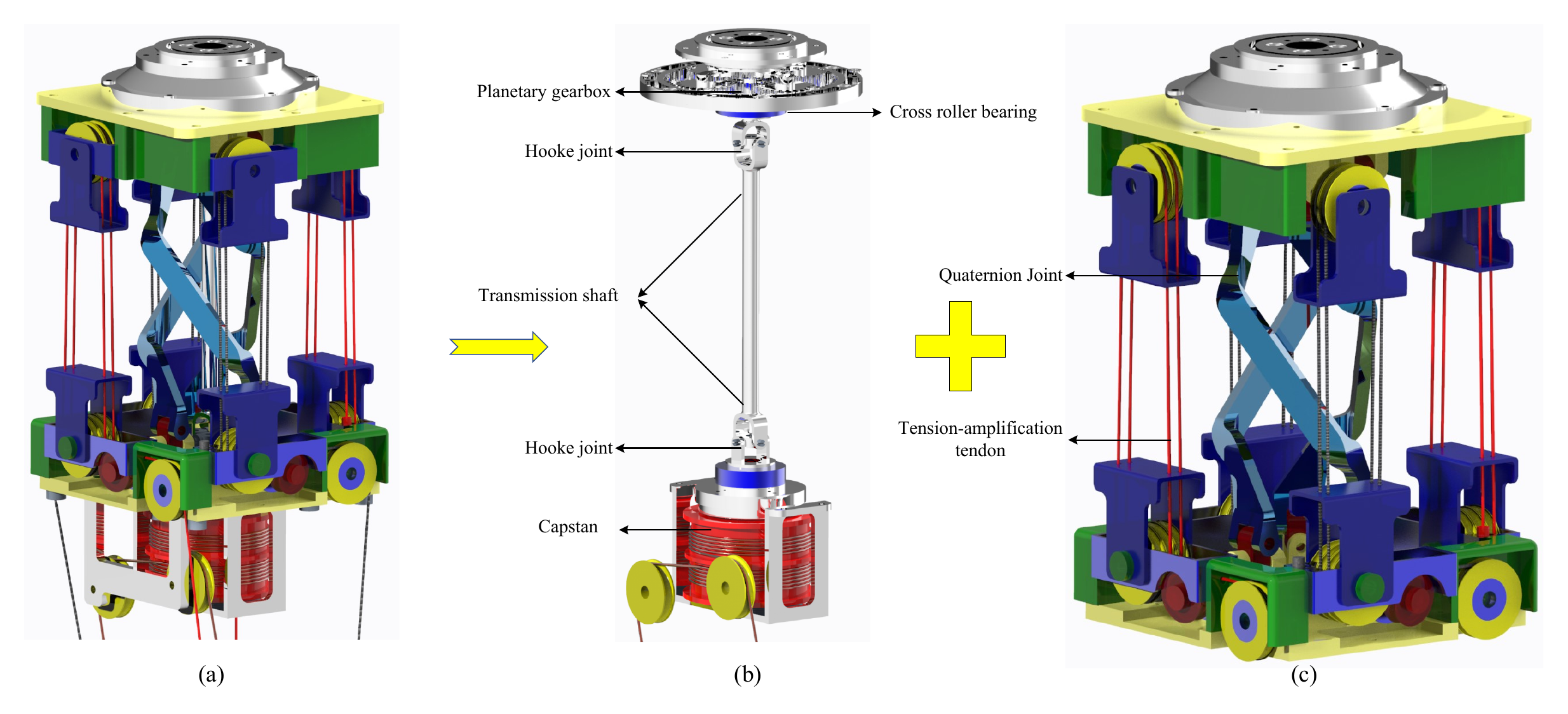}
  \caption{Detail design of the wrist joint.(a) Wrist joint(3-DOF).(b) Rotation Joint(1-DOF).(c)Wrist flexion and extension deflection joints(2-DOF)}
  \label{figure:Detail design of the wrist joint}
  \small\textit{Note:PDF format, 1-column fitting image}
\end{figure}

The design details of the wrist joint are depicted in Fig(\ref{figure:Detail design of the wrist joint}), where the quaternion joint serves as the primary structure of the 2-degree-of-freedom parallel joint in the ATDM wrist, as shown in Fig.(\ref{figure:Detail design of the wrist joint}(c)). Like the antagonistic tendon movement idea in the elbow joint, when the link sizes of each straight link of the quaternion joint meet certain relationships, the pure rolling between the ellipsoids representing the relative movement of its dynamic and static platforms can approximate the pure rolling of two spherical surfaces. In such a configuration, a balanced change in length occurs in a pair of antagonistic tendons, symmetrically driving the joint along its central axis. Consequently, only two motors are required to actuate the four tendons responsible for wrist rotation.

As demonstrated in Fig.(\ref{figure:Detail design of the wrist joint}(b)), for the joint of forearm internal and external rotation, to prevent its movement from coupling with the wrist tendon's flexion and deflection movements, we set this joint last, connected to the end-effector. To prevent wrist movement from affecting the joint tendon movement, the driving tendon of this joint is set inside the forearm before the quaternion joint. The capstan set inside the forearm rotates under the traction of the actuator distributed in the base of the manipulator, and the capstan's rotation is transmitted to the end through the transmission shaft. The transmission shaft passes through two Hooke's joints in series, allowing the shaft to transmit the winch's rotation to the end-effector while complying with the wrist's rolling movement. The issue of unequal speeds at both ends of the universal joint during rotation is also resolved in the process of using paired Hooke's joints in the transmission shaft. The Peek planetary gearbox set at the end can elevate the output torque and stiffness this direction.

\subsubsection{Kinematic Analysis of the Wrist Joint}

\begin{figure}[htbp]
    \centering
    \subfigure[]{
        \includegraphics[width=0.22\textwidth]{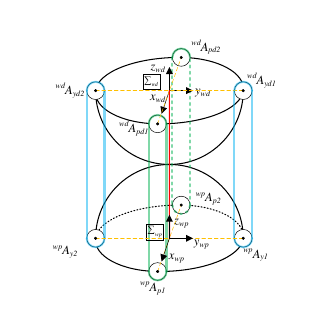}
        \label{fig:Schematic diagram of wrist joint motion a}
    }\hfill
    \subfigure[]{
        \includegraphics[width=0.24\textwidth]{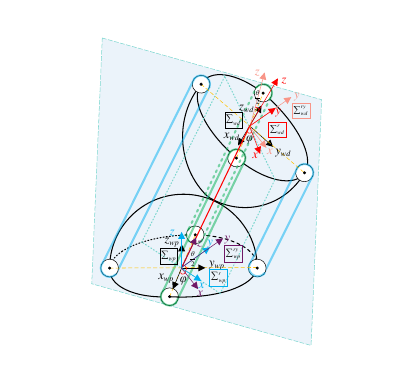}
        \label{fig:Schematic diagram of wrist joint motion b}
    }
    \caption{Schematic diagram of wrist joint motion.(a) Initial state.(b)Bending state}
    \label{figure:Schematic diagram of wrist joint motion}
    \small\textit{Note:PDF format, 1-column fitting image}
\end{figure}

Fig.(\ref{figure:Schematic diagram of wrist joint motion}) illustrates the relationship between the motion of the driving steel cable and the 2-DOF wrist joint. The end of the forearm is relatively fixed, and a base coordinate system $ {\Sigma _{wp}}$ is established at the proximal end. The wrist joint's distal end acts as the moving end ${\Sigma _{wd}}$. The x-axis of both coordinate systems runs along the wrist tendons in the yaw direction, and the y-axis runs along the pitch direction tendons.Similar to section \ref{section:3.2.2},for easier identification and readability,intermediate coordinate systems (${\sum _{wp}} \to \sum _{wp}^r \to \sum _{wp}^{rp} \to \sum _{wd}^{rp} \to \sum _{wd}^r \to {\sum _{wp}}$) are introduced to facilitate the description of the homogeneous transformations between $ {\Sigma _{wp}}$ and ${\Sigma _{wd}}$,colored bounding boxes have been added around the frames of these intermediate coordinate systems which is correspond to the axes of their respective coordinate systems.

Assuming the tendon is a distance $w$ from the center of the wrist, and considering the direction of bending $\phi$ and the angle of bending $\theta$ , the movement of the 2-DOF part of the wrist joint can be simplified as pure rolling on a plane at an angle $\phi$ to the $XOZ$ plane of the proximal coordinate system of the wrist joint. The homogeneous transformation matrix from the distal end to the extreme end of the 2-DOF section, denoted ${}^{wp}\bm{T}_{wd} \in SE(3)$, is
{
\footnotesize
\begin{align}
&^{wp}\bm{T}_{wd}  = Rot(Z,\phi) Rot(Y,\theta/2) Trans(Z,h) Rot(Y,\theta/2) Rot(Z,-\phi)  \\
& = \left[ \begin{array}{cccc}
1 - 2C^2(\phi) S^2(\theta/2) & - S(2\phi) S^2(\theta/2) & C(\phi) S(\theta) & hC(\phi) S(\theta/2) \nonumber\\
- S(2\phi) S^2(\theta/2) & 1 - 2S^2(\phi) S^2(\theta/2) & S(\phi) S(\theta) & hS(\phi) S(\theta/2) \nonumber\\
- C(\phi) S(\theta) & - S(\phi) S(\theta) & C(\theta) & hC(\theta/2) \nonumber\\
0 & 0 & 0 & 1
\end{array} \right]
\end{align}
}

Under such motion, the movement of the tendons in the pitch and roll directions can be calculated as 
{
\begin{align}
L_{p1} &= \left| {}^{wp}{\bm{A}_{p1}} - {}^{wp}{\bm{T}_{wd}}{}^{wd}{\bm{A}_{pd1}} \right|, \notag\\
L_{p2} &= \left| {}^{wp}{\bm{A}_{p2}} - {}^{wp}{\bm{T}_{wd}}{}^{wd}{\bm{A}_{pd2}} \right|, \notag\\
L_{y1} &= \left| {}^{wp}{\bm{A}_{y1}} - {}^{wp}{\bm{T}_{wd}}{}^{wd}{\bm{A}_{yd1}} \right|, \notag\\
L_{y2} &= \left| {}^{wp}{\bm{A}_{y2}} - {}^{wp}{\bm{T}_{wd}}{}^{wd}{\bm{A}_{yd2}} \right|.
\end{align}
}
where, $L_p1$ , $L_p2$, $L_y1$ , and $L_y2$ respectively represent the two sets of tendons in the pitch and yaw directions, where ${}^{wp}{\bm{A}_{p1}} = {}^{wd}{\bm{A}_{pd1}} = \left( {w,0,0,1} \right)$;${}^{wp}{\bm{A}_{p2}} = {}^{wd}{\bm{A}_{pd2}} = \left( { - w,0,0,1} \right)$;${}^{wp}{\bm{A}_{y1}} = {}^{wd}{\bm{A}_{yd1}} = \left( {0,w,0,1} \right)$;${}^{wp}{\bm{A}_{y2}} = {}^{wd}{\bm{A}_{yd2}} = \left( {0, - w,0,1} \right)$;These are the homogeneous coordinates of the endpoints of the TATs. For each pair of mutually coupled motion tendons, the amount of change in their lengths is the same. By substituting the homogeneous coordinates into the equation, we can compute the total change in length of the TATs as
\begin{equation}
\begin{array}{l}
\Delta L_{pitch} = 2N_w w \cos(\varphi) \sin(\theta/2)\\
\Delta L_{yaw} = 2N_w w \sin(\varphi) \sin(\theta/2)
\end{array}\label{Equation:16}
\end{equation}
where, $N_w$ represents the number of windings of the TAT steel cable in the wrist.

The 3-DOF joint at the end of manipulator can flexibly adjust orientation, allowing the AMS to adapt to complex environments while decreasing adjustment-induced disturbances. 

\subsection{Finite Element Analysis}

The finite element analysis (FEA) of the ATDM encompasses strength, stiffness, and topology optimization.

Strength and stiffness were analyzed using \emph{Abaqus 2021}, focusing on material selection and key parameters of the manipulator. For heavier components, boundary conditions were set based on constraints and forces, with topology optimization performed in \emph{Altair Inspire 2022}.


In aerial manipulation, the primary structure of the manipulator must possess high strength and stiffness to counteract substantial forces and torques. Given that aluminum alloys exhibit stiffness an order of magnitude higher than commonly used additive manufacturing plastics, such as PC and ABS, the main structure of the ATDM (encompassing joints and links) was fabricated from aluminum alloy. This choice was made to bolster stiffness, diminish weight, and enhance end positioning accuracy.

For the iterative design, components like gears and pulleys, responsible for transmitting cable tension, don't necessitate the same rigidity as the primary structure. To ensure adaptability in diverse environments, including underwater exploration, fire rescue, and electrical operations, Peek was chosen for these components. Peek is renowned for its wear resistance, stable chemical structure, superior mechanical properties, low friction coefficient, and fatigue resistance, making it apt for cable-driven mechanisms~\cite{wang2021mechanical}.

 Table \ref{table:material} displays the material choices for each component.
\begin{table}[h!]
  \centering
  \caption{Material Selection}
  \label{table:material}
  \begin{tabular}{lc}
    \hline
    Items & Material  \\
    \hline
    joint & Aluminum alloy(7075) \\
    link(Except for the forearm) & Aluminum alloy(7075)  \\
    link(forearm) &AlSi10Mg \\
    gear & Peek \\
    pulley & Peek  \\
    end-effector & Pc  \\
    \hline
  \end{tabular}
\end{table}

Topology optimization, irrespective of initial models and engineering experience, determines optimal structures under constraints, commonly used in aerospace lightweight design.

However, the intricate geometries from topology optimization are hard to realize with traditional manufacturing. This demands a trade-off between technology and optimization, complicating the achievement of optimal product performance. Additive manufacturing, especially Selective Laser Melting (SLM), can accurately produce these optimal structures.

\begin{figure}[htbp]
  \centering
  \includegraphics[width=0.5\textwidth]{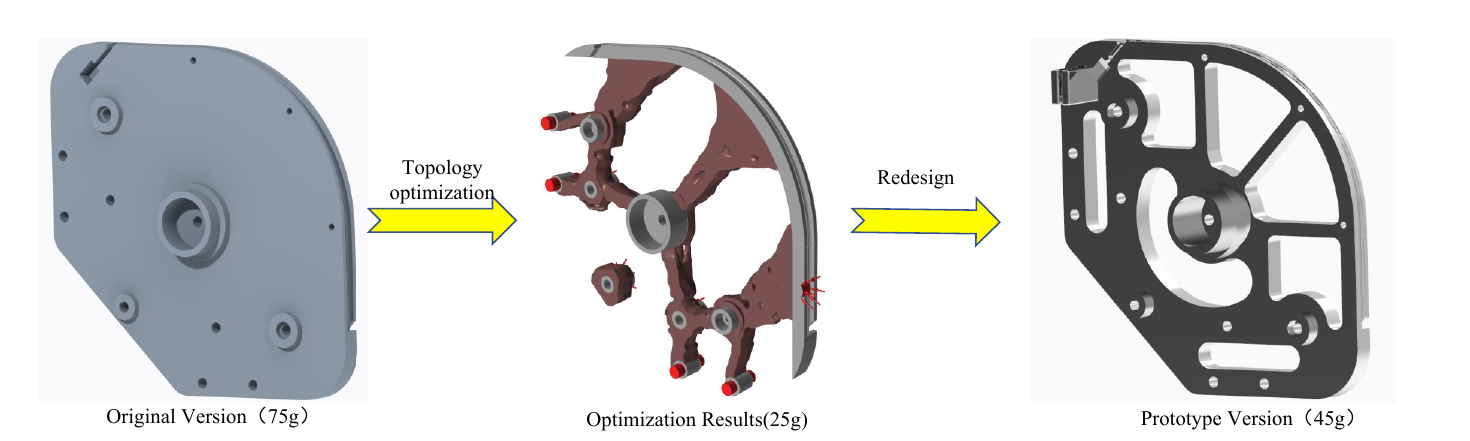}
  \caption{Finite element-based weight reduction of elbow joint}
  \label{figure:Finite element-based weight reduction of elbow joint}
  \small\textit{Note:PDF format, 1-column fitting image}
\end{figure}

Considering the TAT system setup and pulleys for cable routing at each joint, a post-optimization redesign was based on the joint's topology optimization results, as in Fig.(\ref{figure:Finite element-based weight reduction of elbow joint}). The joint's weight was reduced by \(40\% \) from its original based on optimization calculations.

\begin{figure}[h]
  \centering
  \includegraphics[width=0.5\textwidth]{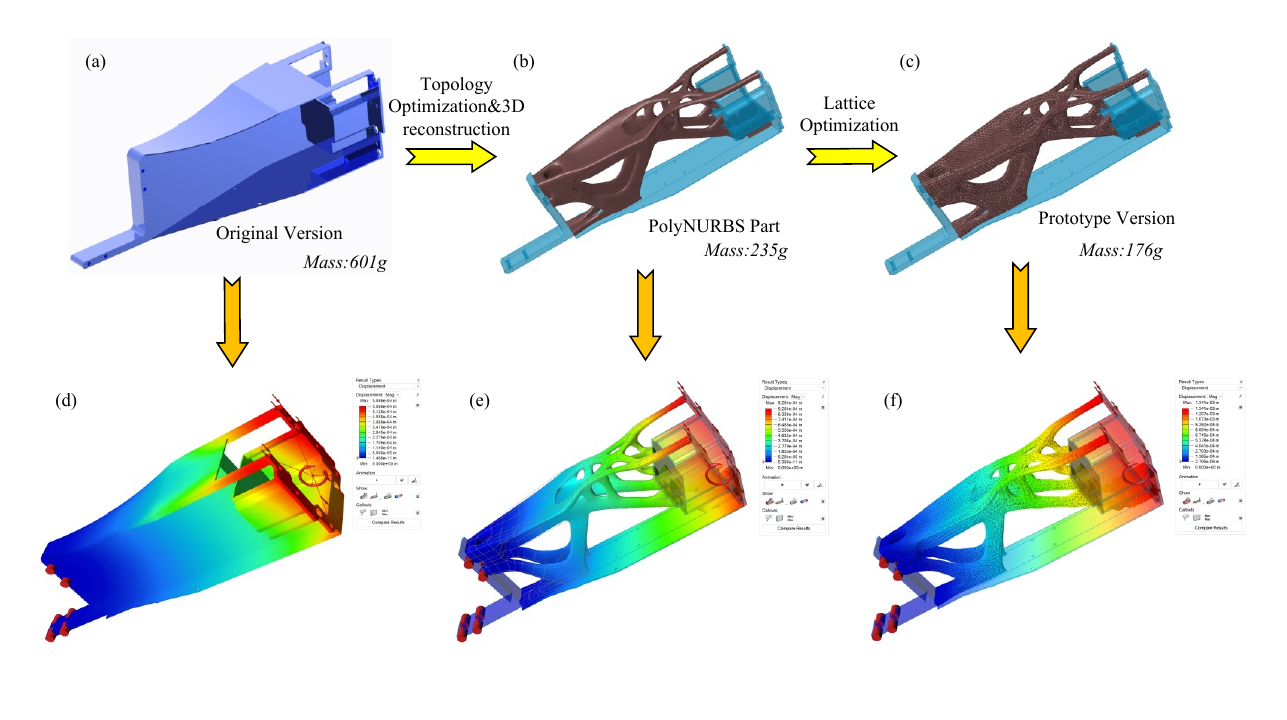}
  \caption{Finite element-based weight reduction of elbow link. (a) Original version. (b) PolyNURBS part. (c) Prototype version. (d) Stiffness analysis of the original. (e) Stiffness analysis of the PolyNURBS part. (f) Stiffness analysis of the Prototype version.}
  \label{figure:Finite element-based weight reduction of elbow link}
  \small\textit{Note:PDF format, 1-column fitting image}
\end{figure}

Aside from the fixed tendon sheath structure of the end-effector's motion path, the links only have to reserve space for installation at the connection with the joint.
In order to obtain better optimization results, this study enlarged the initial design area without interfering with the tendon paths,after setting the design space, topology  optimization calculations were conducted.The calculation maximizes the stiffness under load conditions where multiple combinations of bending and torsional deformations are considered. mass targets are set to 30\% of total design space volume. 

Then,lattice optimization were applied to the links, aiming to mimic the tight radially gradient structure of human bone to reduce weight and inertia while maximizing stiffness under theboundary conditions in which AMS operates.Lattice parameters were set with a target length of \(6\text{mm}\), diameters ranging from \(1\text{mm}\) to \(2\text{mm}\), and filling the design space.

The entire process can refer to Fig.(\ref{figure:Finite element-based weight reduction of elbow link}),Fig.(\ref{figure:Finite element-based weight reduction of elbow link}(a)-(c)) shows the entire optimization process, and Fig.(\ref{figure:Finite element-based weight reduction of elbow link}(d)-(f)) shows the finite element stiffness analysis of the corresponding structure when it is subjected to a large impact (such as AMS falling). Considering the large dynamic load coefficient, a shear force of $500N$ and A bending moment of $15N\cdot m$ is applied to the distal end of the link.

Based on optimization calculations, the weight of the manipulator link was reduced by \(71\% \) from its original weight, which includes the non-optimized area necessary for assembly, and the optimized area is reduced by \(83\% \). Based on the results of finite element stiffness calculation, the stiffness is only increased by 2.4 times, and there is no plastic deformation, which shows that After the elbow link is reduced in weight, it still has a certain residual strength coefficient when facing impact loads on the basis of meeting the strength and stiffness.

\subsection{Overall Configuration of the Manipulator}

\begin{figure*}[t]
  \centering
  \includegraphics[width=1\textwidth]{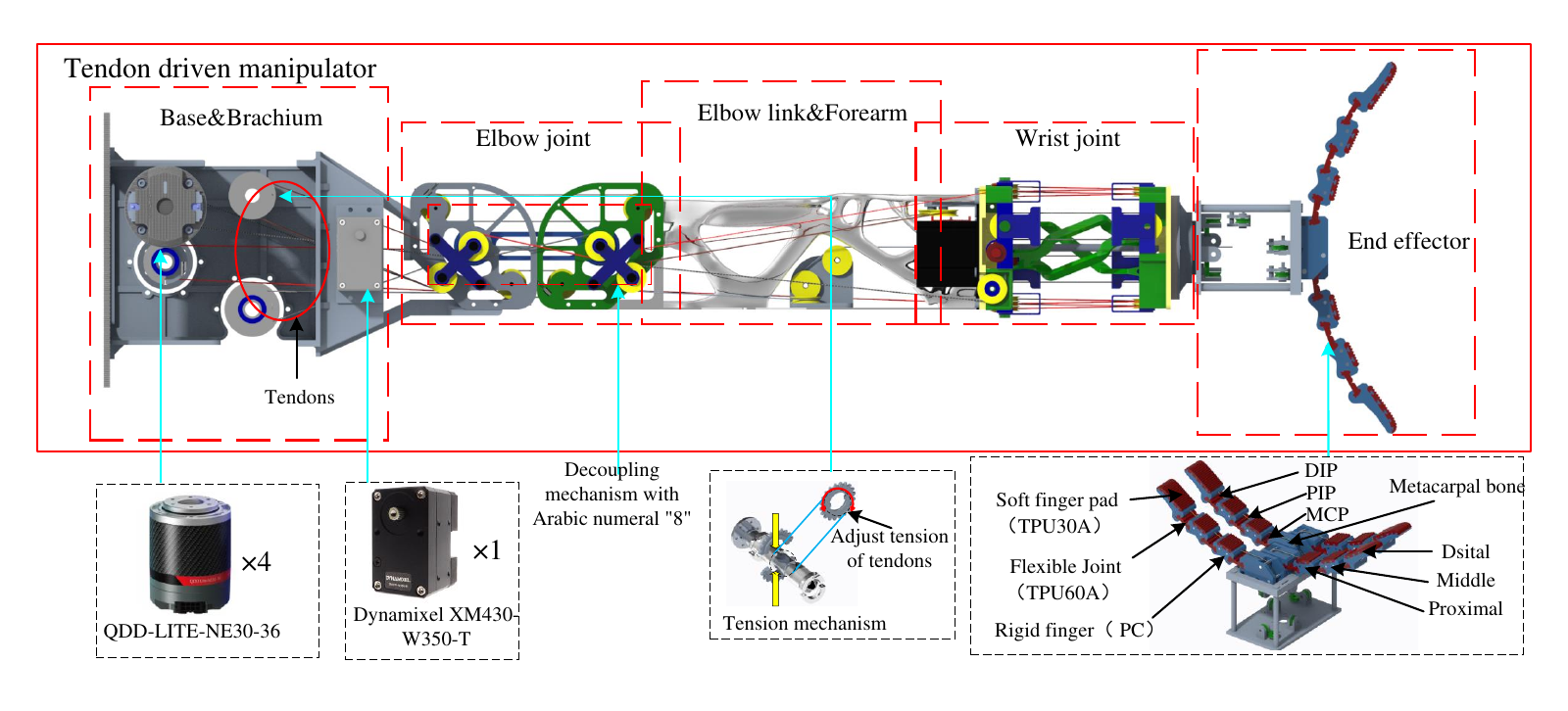}
  \caption{Details of the manipulator}
  \label{figure:Details of the manipulator}
  \small\textit{Note:PDF format, 2-column fitting image}
\end{figure*}

Sections \ref{section:3.2} and \ref{Section:3.3} detail the 1-dof elbow and 3-dof wrist joints. The ATDM comprises a hexrotor and a 4-DOF tendon-driven manipulator with a compliant underactuated soft end-effector. This section showcases the drive, tensioning, and end-effector of the manipulator. Details are shown in Fig.(\ref{figure:Details of the manipulator}).

Actuators are distributed on a UAV-attached base. The design ensures consistent length changes between primary and antagonist tendons during single joint motion. The '8' decoupling pulley on the elbow joint, combined with the pre-elbow link, decouples elbow and wrist joint motions. This setup achieves mutual cancellation of radial forces on the pulley shaft, eliminating the need for additional cable tension retention once each manipulator cable is pretensioned.

Though no tension retention mechanism is needed, mechanical pre-tightening of each steel cableis essential for motion accuracy under load and quick reverse movements. The pre-tensioning force is set to half the steel cable's maximum load.

Considering AMS's weight sensitivity, the pre-tensioning mechanism's weight and size are crucial. While both rotating screw and ratchet methods at the cable end are lightweight, they have issues with force control and smoothness, respectively. The motor-connected winch drum was divided into a rotating shaft, primary muscle drum, and antagonist muscle drum. The shaft connects to the motor, with the muscle capstans tensioned via a sprocket.

Grappling with object manipulation in unstructured settings remains a significant challenge in robotics \cite{dollar2010highly}. While many studies favor two-fingered parallel grippers \cite{fanni2017new, dong2021centimeter}, their dexterity is limited in dynamic environments. In AMS, high-DOF fingers have proven to enhance grasping success due to vibrations during flight. Soft underactuated end-effectors, which passively adapt to object shapes, offer a more efficient approach than traditional manipulators requiring multiple servo motors\cite{hussain2021compliant}. They utilize fewer actuators for versatile grasps \cite{dollar2010highly,hussain2021compliant,ruotolo2021grasping}. Such end-effectors not only enhance AMS versatility but also reduce disturbances from aerial platforms, providing a cost-effective solution.
This paper provides a brief overview of the end-effector's design, as depicted in the lower right part of Fig.(\ref{figure:Details of the manipulator}).

\section{Analysis of the ATDM}\label{section:4}

This section is dedicated to analyzing the workspace of the ATDM and deriving the output torque and joint stiffness of the joints, which are based on the TAT mechanism.

\subsection{Analysis of ATDM's Working Space}

UAV with six degrees of freedom is an underactuated system, there are coupling between its position and attitude loops,
with only four control channels available. The ATDM features a 4-DOF manipulator system. Its 3-DOF joint end allows flexible posture adjustments, minimizing UAV coupling disturbance. Additionally, redundancy in the end's degree of freedom enables tasks like UAV obstacle avoidance, singularity avoidance, joint limit prevention, load optimization, and posture optimization.

Fig. \ref{figure:coordinate_frame} shows the ATDM's schematic with coordinate systems. ${\Sigma _I}\{ {O_I} - {X_I}{Y_I}{Z_I}\}$ denotes the inertial frame, ${\Sigma _B}\{ {O_B} - {X_B}{Y_B}{Z_B}\}$ represents the body coordinate at the UAV's COM, and ${\Sigma _E}\{ {O_E} - {X_E}{Y_E}{Z_E}\}$ is the end-effector's system.

The manipulator is modeled as a 7-DOF arm with three pairs of coupled joints. Each joint's coordinate system is established following section \ref{section:3}, with the MD-H parameters shown in Table \ref{table:MDH}. 

\begin{table}[h!]
  \centering
  \caption{MD-H Parameters of The Manipulator}
  \label{table:MDH}
  \begin{tabular}{lcccc}
    \hline
    Link$_i$ & $\theta_i$ (°) & $d_i$ (mm) & $a_{i-1}$ (mm) & $\alpha_{i-1}$ (°) \\
    \hline
    Link1 & ${\overline\theta}_1$ & 0 & 200 & 0 \\
    Link2 & ${\overline\theta}_1 + \pi/2$ & 0 & 80 & 0 \\
    Link3 & ${\overline\theta}_2$ & 300 & 1 & $\pi/2$ \\
    Link4 & ${\overline\theta}_3 - \pi/2$ & 0 & 0 & $-\pi/2$ \\
    Link5 & ${\overline\theta}_3 + \pi/2$ & 0 & 90 & 0 \\
    Link6 & ${\overline\theta}_2$ & 0 & 0 & $\pi/2$ \\
    Link7 & ${\overline\theta}_4$ & 100 & 0 & 0 \\
    \hline
  \end{tabular}
\end{table}

Due to the existence of virtual coupled joints,there are only four independent angles${\bar \theta _i}(i = 1,2,3,4)$, the ranges of which are as follows: ${\theta _1} \in [ - \pi /4,\pi /4],{\theta _2} \in [ - \pi /4,\pi /4],{\theta _3} \in [ - \pi /4,\pi /4],{\theta _4} \in [ - \pi ,\pi ]$. The description of ${\Sigma _E}$ in ${\Sigma _B}$ can be obtained through the homogeneous transformation matrices between joints, which is represented as ${}^B{\bm{T}_E}$, and the homogeneous transformation matrices ${}^{i - 1}{\bm{T}_i}$ between adjacent links refer to \cite{1986Introduction}.

\begin{figure}[h]
    \centering
    \includegraphics[width=0.5\textwidth]{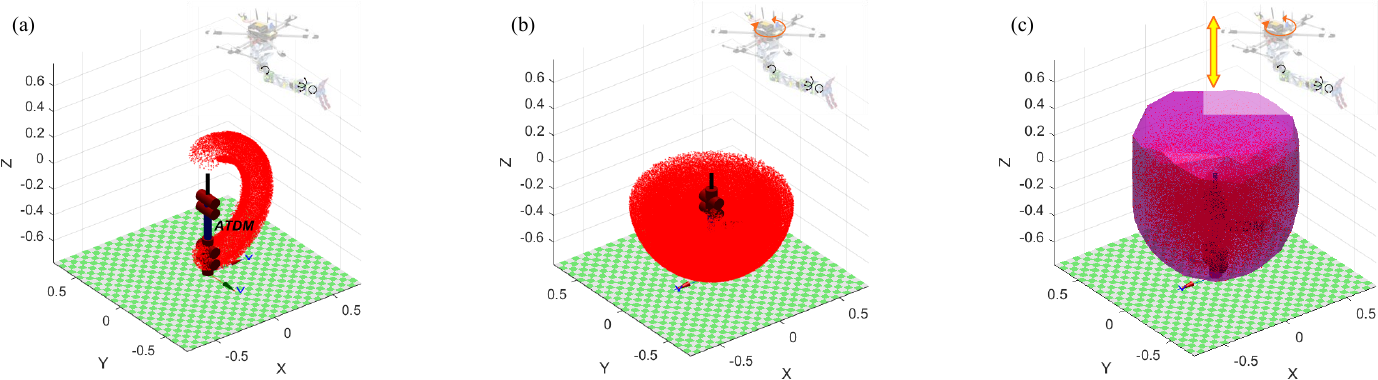}
    \caption{Analysis of ATDM's working space. (a) Hover. (b) Yaw. (c) Vertically ascend and descend}
    \label{figure:workspace}
    \small\textit{Note:PDF format, 1-column fitting image}
\end{figure}

Equipping a UAV with a 4-DOF manipulator enables the end-effector to reach any desired spatial position with any arbitrary orientation. Given the UAV's lack of coupling in gravitational or yaw directions and the typical front-mounted camera, we analyze the ATDM's end-effector workspace under three conditions:
\begin{enumerate}
    \item the UAV remains in hover (Fig. \ref{figure:workspace}(a)),
    \item the UAV hovers at a fixed point with a varying yaw angle (Fig. \ref{figure:workspace}(b)),
    \item the UAV can move along the gravitational axis while also executing yaw movements (Fig. \ref{figure:workspace}(c)).
\end{enumerate}
Assuming the UAV is near its target and needs only pose fine-tuning, we consider only two motion directions. The 3-DOF wrist joint allows flexible attitude adjustments of the end, with minimal position fluctuations, reducing UAV disturbances. In hover mode, the end's workspace is limited. However, with the ATDM system combining a UAV and tendon-driven manipulator, the workspace expands in the latter two scenarios. While the workspace is initially small, it expands significantly when the UAV utilizes its two uncoupled degrees of freedom. This heart-shaped three-dimensional workspace allows for flexible operations.

\subsection{Torque and Stiffness Analysis of ATDM}

This section focuses on analyzing the output torque and stiffness of the elbow and wrist joints in the ATDM.

\subsubsection{Analysis of the 1-DOF Elbow Joint}\label{section:Analysis of the 1-DOF Elbow Joint}

The torque in the joints of the manipulator can be represented as
\begin{equation}
    {\tau _{elbow}}\Delta \theta  = \Delta T\Delta l
\end{equation}

Where $\Delta \theta$ is the joint rotation angle, $\Delta T$ is the difference in tension between the agonist and antagonist muscles in a single cable, and $\Delta l$ is the displacement of the tendon cable. Due to the uniformity of the agonist and antagonist muscles, according to Equ (\ref{equation:3})
\begin{equation}
    \frac{\Delta l}{\Delta \theta} = \frac{N_e d_e}{2} \cos(\theta/2) \label{equation:19}
\end{equation}

The output torque of the elbow joint can be represented as:

\begin{equation}
    \tau_{elbow} = \frac{N_e d_e}{2} \cos(\theta/2) \Delta T \quad s.t \quad \Delta T = \min \left( k_{cable}, \frac{\tau_{motor}^{elbow}}{R_{capstan}} \right) \label{equation:20}
\end{equation}

Where, $k_{cable}$ represents the maximum tensile strength of the steel cable, $\tau _{motor}^{elbow}$ denoting the motor's output torque at this time.

The stiffness of the manipulator is derived as follows:

Assuming the mechanical arm is in equilibrium at angle $\theta$, and a torque $\delta_{\tau}$ is applied to the joint, the joint undergoes a minor deformation $\delta_{\theta}$. According to the principle of virtual work 
\begin{equation}
    \delta {W_{int}} = \delta {W_{out}}
\end{equation}
we obtain
\begin{equation}
    \delta \tau \delta \theta  = \Sigma _{i = 1}^n\delta {T_i}\delta {L_i}\label{equation:22}
\end{equation}

According to Equ.(\ref{equation:19}) and Equ.(\ref{equation:22}),it can be deduced that
{
\begin{align}
    \delta \tau &= \frac{N_e d_e}{2}\cos(\theta/2)(T_{ago} - T_{ant}) \notag \\
    &= \frac{N_e d_e}{2}\cos(\theta/2)\left(k_{cable}\frac{\delta l}{2R - d_e \sin(\theta/2)} + k_{cable}\frac{\delta l}{2R + d_e \sin(\theta/2)}\right) \notag \\
    &= \frac{N_e d_e}{2}\cos(\theta/2)k_{cable}\frac{4R\delta l}{4R^2 - d_e^2 \sin^2(\theta/2)}
\end{align}
}

the stiffness of the elbow joint can be derived as
{
\begin{equation}
    k_{elbow} = \frac{\delta \tau}{\delta \theta} = \frac{N_e d_e}{2} \cos(\theta/2) k_{cable} \frac{4R}{4R^2 - d_e^2 \sin^2(\theta/2)} \frac{\delta l}{\delta \theta} \label{equation:24}
\end{equation}
}

From Equ.(\ref{equation:19}) and Equ.(\ref{equation:24}),$k_{elbow}$ can be calculated as
{
\begin{equation}
    k_{elbow} = \left( N_e d_e \cos(\theta/2) \right)^2 k_{cable} \frac{R}{4R^2 - d_e^2 \sin^2(\theta/2)} \label{equation:25}
\end{equation}
}

According Equ.(\ref{equation:20}) and Equ.(\ref{equation:25}), it's clear that increasing $N_e$ and $d_e$ can obtain higher joint torque. In the design of the elbow joint, we have chosen $R=45mm$,$N_e=6$, ${d_e} = 48\sqrt 2mm$ and $R_{capstan }=11mm$,These are designable parameters.

\subsubsection{Analysis of the 3-dof Wrist Joint}

The pitch and yaw motions of the 2-DOF portion of the wrist joint can be calculated using Equ(\ref{Equation:16}) and the calculations for the 1-DOF joint in section(\ref{section:Analysis of the 1-DOF Elbow Joint}). The resultant output torque and stiffness for a single pair of antagonistic tendons can be obtained likewise. When both tendons are engaged, the total output torque for the 2-dof section is as follows:
{
\begin{align}
{\tau _{wrist - 2dof}} &= {N_w}w\cos (\theta/2)(\Delta {T_{pitch}}\left| {\sin (\varphi )} \right| + \Delta {T_{yaw}}\left| {\cos (\varphi )} \right|) \nonumber\\
\Delta {T_{pitch}} &= \min ({k_{cable}},\frac{{\tau _{motor}^{pitch}}}{{{R_{capstan}}}}) \nonumber\\
\Delta {T_{yaw}} &= \min ({k_{cable}},\frac{{\tau _{motor}^{yaw}}}{{{R_{capstan}}}})
\label{equation:26}
\end{align}
}
Similarly, the joint stiffness for the 2-dof part when both tendons are engaged can be calculated as
{
\begin{align}
    k_{wrist - 2dof} &= \frac{1}{2} \left( 2N_w w \cos(\theta/2) \cos(\varphi) \right)^2 k_{cable} \nonumber\\
    &+ \frac{1}{2} \left( 2N_w w \cos(\theta/2) \sin(\varphi) \right)^2 k_{cable} \nonumber\\
    &= 2N_w^2 w^2 \cos^2(\theta/2) k_{cable}
\label{equation:27}
\end{align}
}
A capstan with a radius of $R_{capstan }^{roll}$ is used to control the motion of the wrist joint in the roll direction, rather than to amplify tendon tension. Likewise, we can calculate the stiffness and output torque based on a planetary gear reducer with a gear ratio $N_r$ as
{
\begin{equation}
\begin{split}
{\tau _{wrist - roll}} &= {N_r}R_{caps\tan }^{roll}\Delta T;{\rm{ }}s.t{\rm{ }}\Delta T = \min ({k_{cable}},\frac{{\tau _{motor}^{elbow}}}{{{R_{caps\tan }}}})\\
{k_{wrist - roll}} &= 2{\left( {{N_r}R_{caps\tan }^{roll}} \right)^2}{k_{cable}}
\end{split}
\label{equation:28}
\end{equation}
}
From Equ.(\ref{equation:26}) to Equ.(\ref{equation:28}), it's evident that increasing \( N_w \) and \( w \) boosts the 2-DOF wrist joint's output torque and stiffness. Increasing the gear ratio \( N_r \) improves the output torque and stiffness in the roll direction. Given the requirements for aerial grasping, the torque needed in pitch and yaw motions significantly exceeds that in the roll direction. In the design of the elbow joint, we have chosen \( N_w=4 \), \( w=40mm \), \( N_r=5 \), \( R_{capstan}=11mm \), and \( R_{capstan}^{roll}=25mm \).

Referring to Equ.(\ref{equation:28}), with \( N_r=1 \), the output torque and stiffness expressions for the traditional cable-roller driven manipulator can be similarly derived. By comparing with Equ.(\ref{equation:20}),(\ref{equation:25}),(\ref{equation:26}),(\ref{equation:27}), it is clear that the TAT mechanism effectively increases the output torque and output stiffness.

\section{Simulation Analysis}\label{section:5}

\begin{figure}[htbp]
  \centering
  \includegraphics[width=0.3\textwidth]{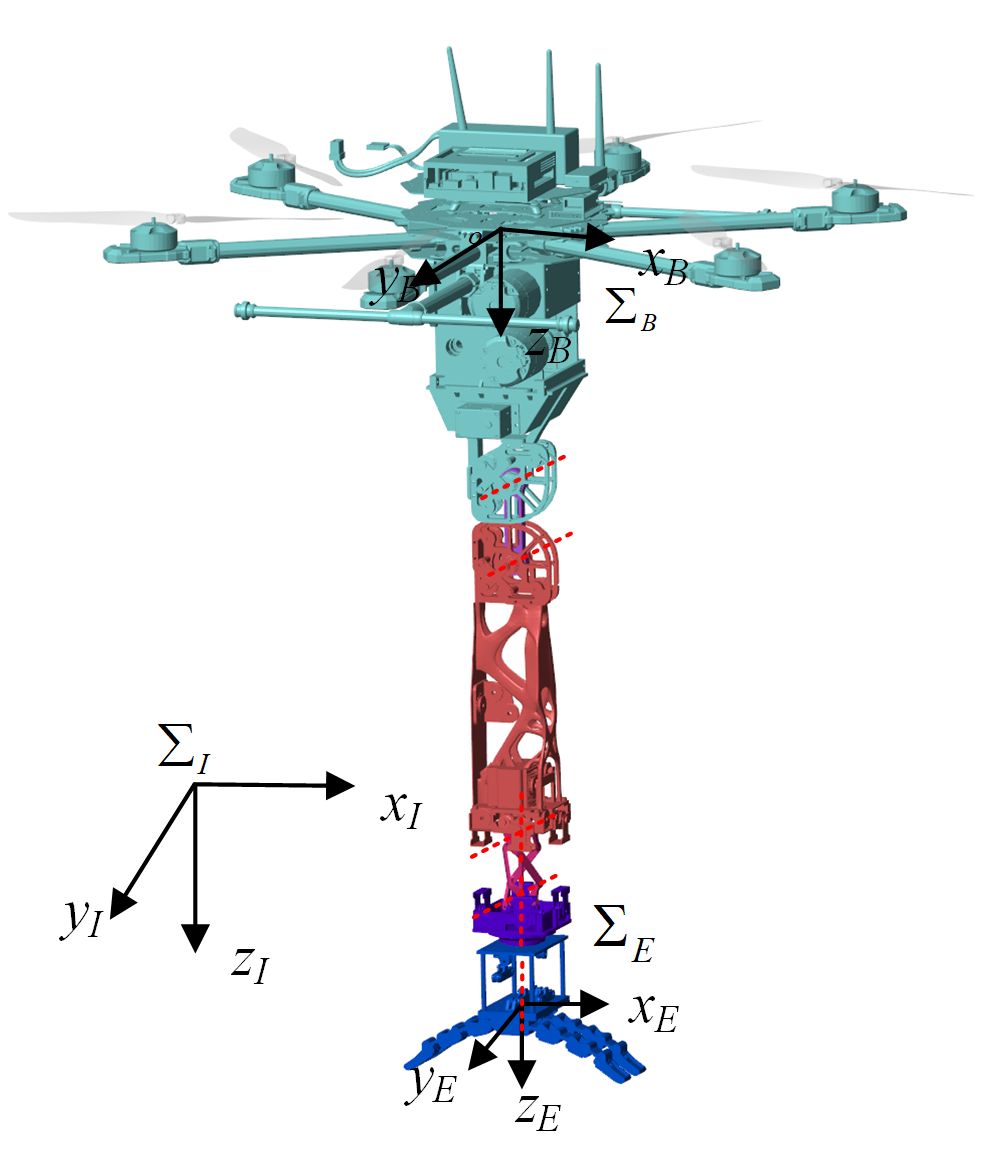}
  \caption{Semi-physical simulation of ATDM}
  \label{figure:Semi-physical simulation of ATDM}
  \small\textit{Note:JPG format, 1-column fitting image}
\end{figure}

{
\footnotesize
\begin{table}[h!]
  \centering
  \caption{Mass \& Inertia of UAV and Each Link of ATDM}
  \label{mass&inertial-table}
  \small
  \begin{tabular}{lcc}
    \hline
    Items & Mass $\left( \text{kg} \right)$ & Inertial $\left( \text{kg} \cdot \text{m}^2 \right)$ \\
    \hline
    UAV & 2.711 & $\text{diag}\left(1.1 \cdot 10^{-1}, 1.1 \cdot 10^{-1}, 2.2 \cdot 10^{-1}\right)$ \\
    Base & 1.899 & $\text{diag}\left(1.1 \cdot 10^{-2}, 1.6 \cdot 10^{-2}, 8.5 \cdot 10^{-3}\right)$ \\
    Joint1 & $2.35 \cdot 10^{-2}$ & $\text{diag}\left(2.2 \cdot 10^{-5}, 2.4 \cdot 10^{-5}, 1.5 \cdot 10^{-6}\right)$ \\
    Joint2 & $5.36 \cdot 10^{-1}$ & $\text{diag}\left(6.2 \cdot 10^{-3}, 6.4 \cdot 10^{-3}, 5.1 \cdot 10^{-4}\right)$ \\
    Joint3 & $1.11 \cdot 10^{-3}$ & $\text{diag}\left(1.7 \cdot 10^{-9}, 1.7 \cdot 10^{-9}, 2.9 \cdot 10^{-9}\right)$ \\
    Joint4 & $2.4 \cdot 10^{-2}$ & $\text{diag}\left(1.7 \cdot 10^{-5}, 1.7 \cdot 10^{-5}, 4.3 \cdot 10^{-6}\right)$ \\
    Joint5 & $1.11 \cdot 10^{-3}$ & $\text{diag}\left(1.7 \cdot 10^{-9}, 1.7 \cdot 10^{-9}, 2.9 \cdot 10^{-9}\right)$ \\
    Joint6 & $6.26 \cdot 10^{-2}$ & $\text{diag}\left(6.7 \cdot 10^{-5}, 6.7 \cdot 10^{-5}, 1.0 \cdot 10^{-4}\right)$ \\
    Joint7 & $1.70 \cdot 10^{-1}$ & $\text{diag}\left(3.0 \cdot 10^{-4}, 6.3 \cdot 10^{-4}, 5.8 \cdot 10^{-4}\right)$ \\
    QDD-NE30-36 & $2.5 \cdot 10^{-1}$ & $\text{diag}\left(1.7 \cdot 10^{-4}, 1.7 \cdot 10^{-4}, 1.2 \cdot 10^{-4}\right)$ \\
    XM430-W350 & $8.0 \cdot 10^{-2}$ & $\text{diag}\left(2.2 \cdot 10^{-5}, 1.3 \cdot 10^{-5}, 1.9 \cdot 10^{-5}\right)$ \\
    \hline
  \end{tabular}
\end{table}
}

To validate the effectiveness of the proposed ATDM in decreasing coupling disturbance during motion compared to traditional serial manipulators, we established a semi-physical simulation environment using Simscape, which provides  facilitates multi-body dynamic simulations for the ATDM, as presented in Fig. (\ref{figure:Semi-physical simulation of ATDM}).

\begin{figure}[h!]
    \centering
    \subfigure[]{
        \includegraphics[width=0.23\textwidth]{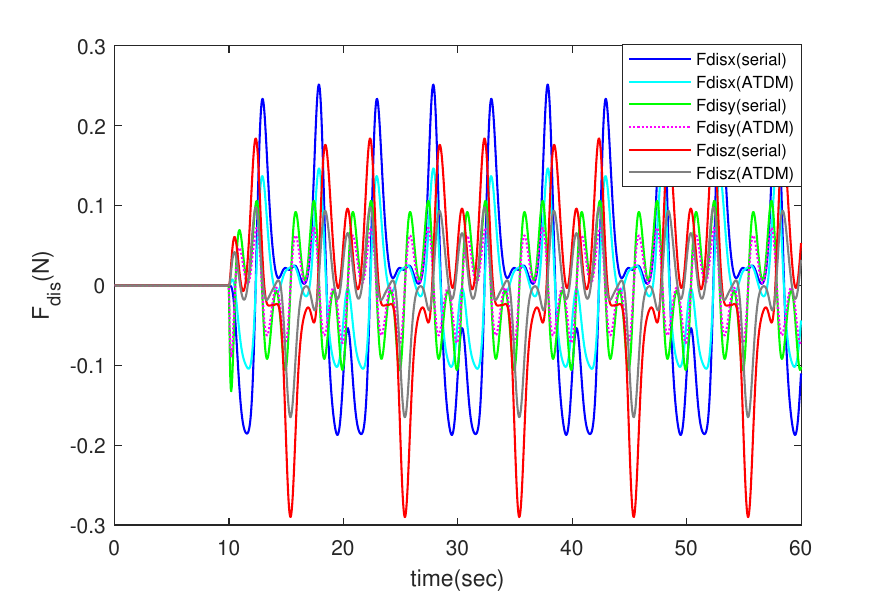}
        \label{fig:simulation a}
    }\hfill
    \subfigure[]{
        \includegraphics[width=0.23\textwidth]{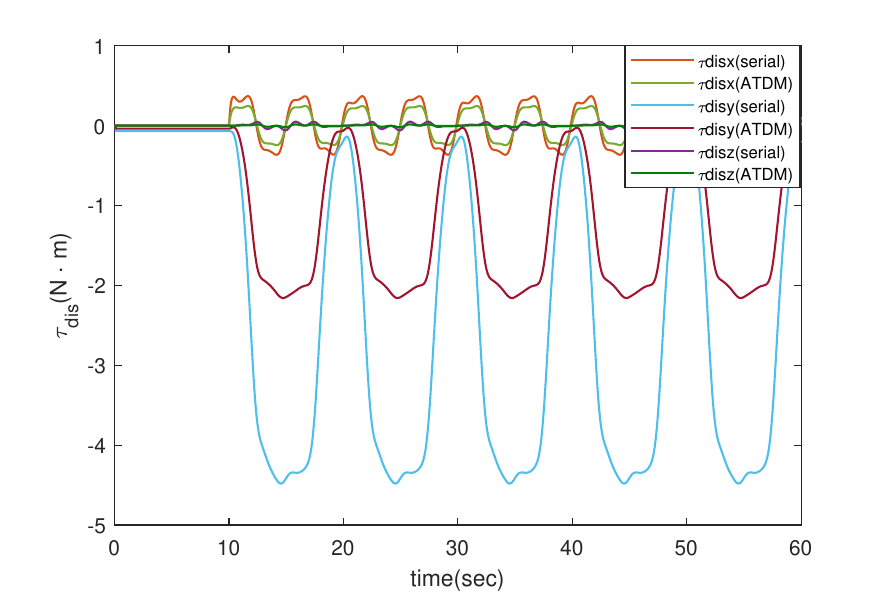}
        \label{fig:simulation b}
    }\\
    \subfigure[]{
        \includegraphics[width=0.23\textwidth]{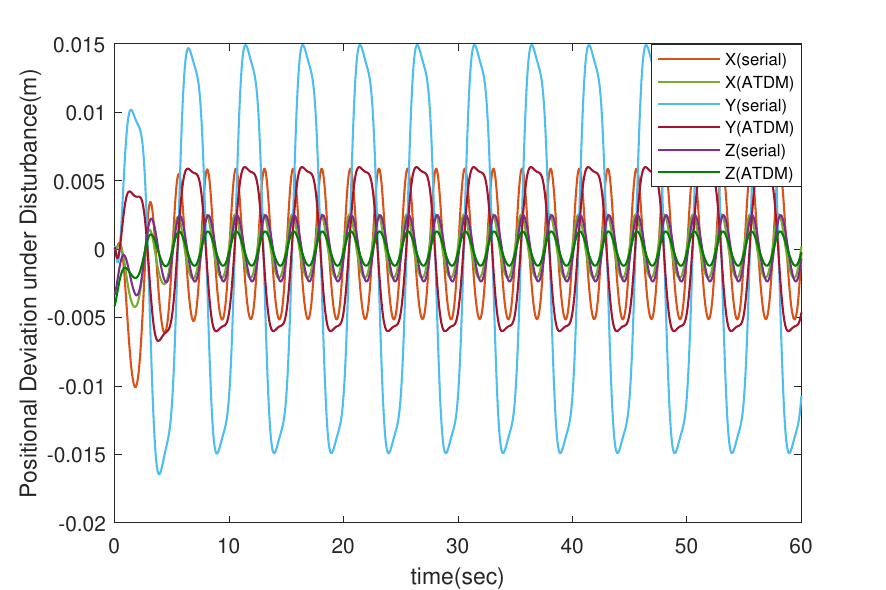}
        \label{fig:simulation c}
    }\hfill
    \subfigure[]{
        \includegraphics[width=0.23\textwidth]{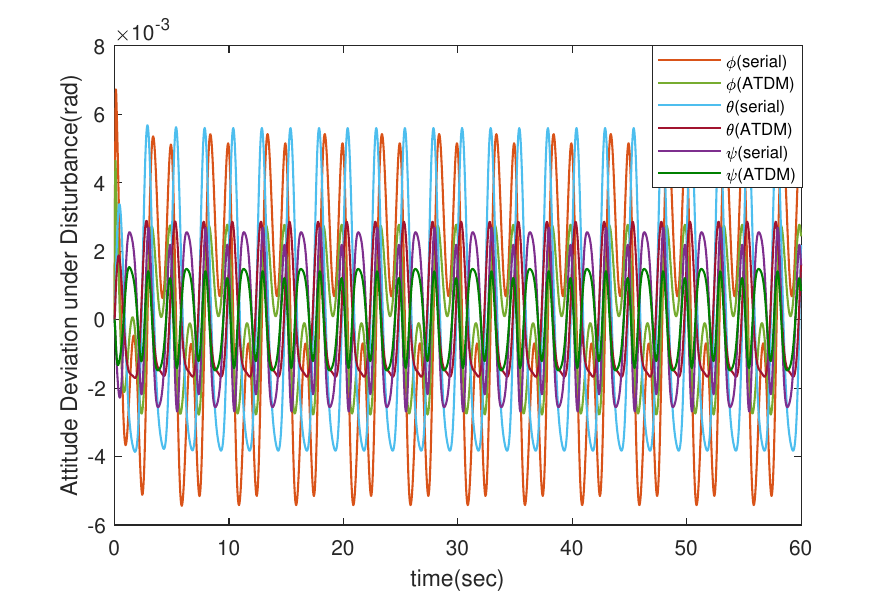}
        \label{fig:simulation d}
    }\\
    \subfigure[]{
        \includegraphics[width=0.23\textwidth]{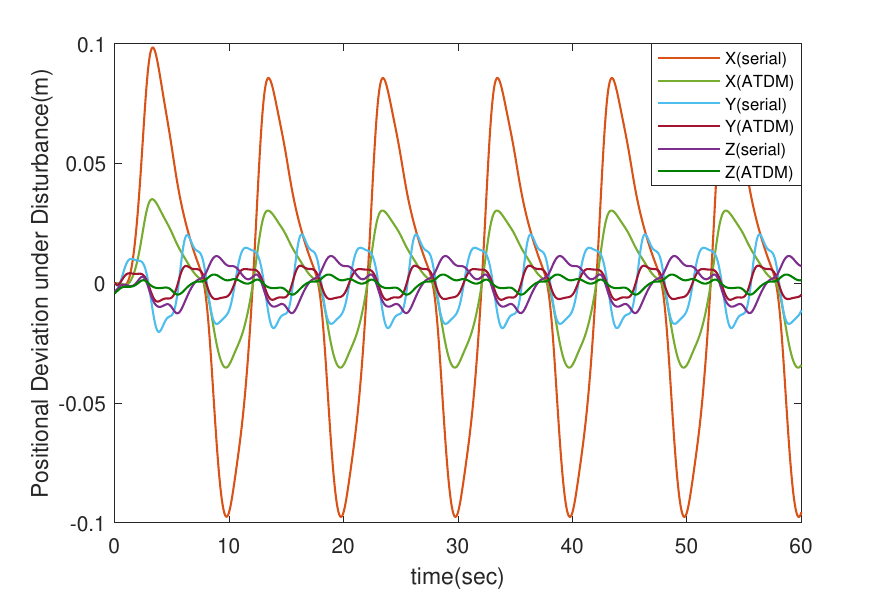}
        \label{fig:simulation e}
    }\hfill
    \subfigure[]{
        \includegraphics[width=0.23\textwidth]{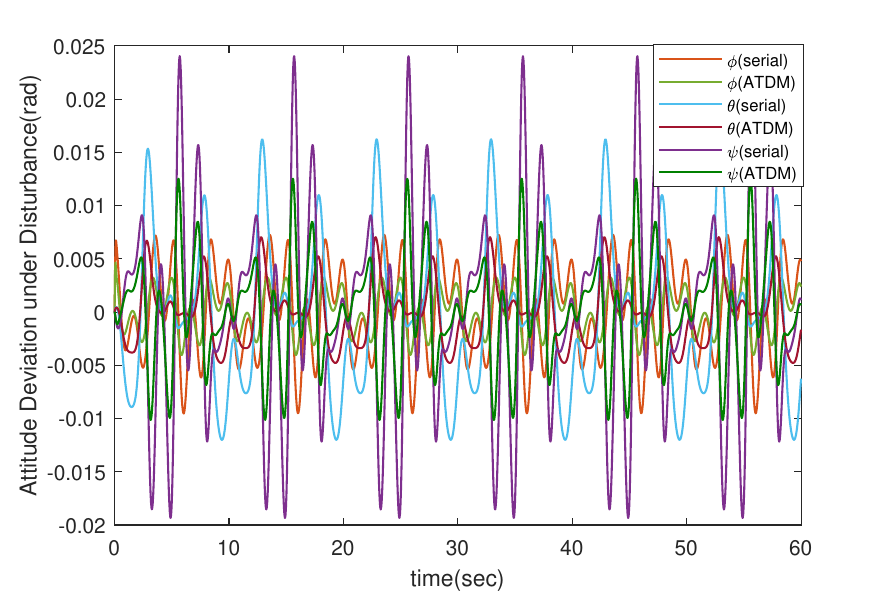}
        \label{fig:simulation f}
    }
    \caption{Simulation result.(a)Force of coupling disturbance under wide-range of rapid movement.(b)Torque of coupling disturbance under wide-range of rapid movement.(c)(c)Position deviation of UAV under wrist joint motion.(d)Attitude deviation of UAV under wrist joint motion.(e)Position deviation of UAV under wide-range of rapid movement.(f)Attitude deviation of UAV under wide-range of rapid movement}
    \label{fig:simulation}
    \small\textit{Note:PDF format, 1-column fitting image}
\end{figure}

For a comparative analysis with the serial manipulator, the mass and inertia of actuators for each joint and the end-effector are allocated at the COM of the preceding link. This is a conservative approach in contrast to allocations at the joints. The mass, length, and inertia parameters for the AMS are outlined in Table \ref{mass&inertial-table}. Where the parameters of the base have taken into account the parameters of the actuators of both manipulator and end-effector.Simulations were performed under three working conditions. 

\subsection{Coupling Disturbance Forces and Torques}

Using the coupling disturbance model from section \ref{subsection:2.2}, we compare the disturbances on the UAV from both manipulator types. Under this condition, once the AMS reaches the target location, the UAV's movement is restricted. The manipulator's 1-DOF elbow joint performs sinusoidal oscillations with an amplitude of \(\pi/2\) and a period of 5 seconds, while the wrist joint angles also oscillate sinusoidally with an amplitude of \(\pi/2\) and periods of 10 seconds. The results of these simulations are depicted in Fig.(\ref{fig:simulation a}) and Fig.(\ref{fig:simulation b}). 

To evaluate the magnitude of the disturbance force and moment during the simulation time, we introduce two metrics: mean absolute variable (MAV) and root mean square variable (RMSV). The expressions for these metrics are:
{
\begin{equation}
    \text{MAV} = \frac{1}{n} \sum_{i=1}^{n} \left| y_i  \right|
\end{equation}

\begin{equation}
    \text{RMSV} = \sqrt{\frac{1}{n} \sum_{i=1}^{n} (y_i )^2}
\end{equation}
}
Table \ref{coupling disturbance forces and torques-table} presents a comparison of the coupling disturbance for both manipulator types, using MAV and RMSV as metrics.

\begin{table}[h!]
  \centering
  \caption{Comparison of coupling disturbance forces and torques}
  \label{coupling disturbance forces and torques-table}
  \begin{tabular}{lcc}
    \hline
    Items & MAV:ATDM/serial & RMSV:ATDM/serial \\
    \hline
    \( F_{\text{dis}} \) & 0.5548 & 0.5591 \\
    \( \tau_{\text{dis}} \) & 0.4843 & 0.4757 \\
    \hline
  \end{tabular}
\end{table}

\subsection{Flexible Wrist Movement of Manipulator in AMS Hovering State}

In this scenario, once the AMS reaches its target position, it remains in a hovering state. The manipulator's 3-DOF wrist joint is subjected to sinusoidal oscillations, with each joint oscillating with an amplitude of \(\pi/2\) and a period of 10 seconds. 

To analyze the effects of different manipulators on the AMS system, a robust controller, as detailed in \cite{cao2023eso}, is utilized. The outcomes of these simulations are illustrated in Figures \ref{fig:simulation c} and \ref{fig:simulation d}.

Consistent with the previous section, we employ the same metrics to evaluate the impact of the two manipulator types on the AMS's positional and attitudinal stability during fixed-point hovering. The results are summarized in Table \ref{pose change with wrist motion-table}.

\begin{table}[h!]
  \centering
  \caption{Comparison of pose changes during AMS hovering with wrist motion}
  \label{pose change with wrist motion-table}
  \begin{tabular}{lcc}
    \hline
    Items & MAV:ATDM/serial & RMSV:ATDM/serial \\
    \hline
    Position & 0.4244 & 0.4215 \\
    Attitude & 0.5017 & 0.5044 \\
    \hline
  \end{tabular}
\end{table}

\subsection{Wide-Range Rapid Movement of Manipulator in AMS Hovering State}

In this scenario, utilizing the same controller as in the second working condition, the AMS, upon reaching its desired position, remains in a hovering state. The manipulator operates similarly to the first working condition. The results of these simulations are depicted in Figures \ref{fig:simulation e} and \ref{fig:simulation f}.

\begin{table}[h!]
  \centering
  \caption{Comparison of pose changes during AMS hovering with wide-range rapid manipulator movement}
  \label{pose change with wide-range of rapid movement-table}
  \begin{tabular}{lcc}
    \hline
    Items & MAV:ATDM/serial & RMSV:ATDM/serial \\
    \hline
    Position & 0.3738 & 0.3644 \\
    Attitude & 0.4760 & 0.4884 \\
    \hline
  \end{tabular}
\end{table}

Consistent with the previous sections, we evaluate the impact of both the ATDM and the serial manipulator on the AMS's stability. The results are summarized in Table \ref{pose change with wide-range of rapid movement-table}.

\subsection{Summary and Conclusions}

Upon reviewing Tables \ref{coupling disturbance forces and torques-table}, \ref{pose change with wrist motion-table}, and \ref{pose change with wide-range of rapid movement-table}, it becomes clear that modifying the actuator distribution in the tendon-driven manipulator can significantly reduce the coupling disturbance within the AMS system. This reduction is particularly evident during high-speed, wide-range movements of the manipulator. It is noteworthy that in the semi-physical simulation system, a rather conservative approach was adopted by assigning the physical properties of the actuator to the center of mass of the preceding link. This inevitably results in less coupling disturbance compared to the real-world scenario where the actuator is located at the joint in a serial manipulator. This further corroborates the significant improvement demonstrated by the proposed ATDM system in minimizing coupling interference, as opposed to the traditional AMS system.

\section{Conclusion}\label{section:6}

To address the challenges of load, stiffness, and coupling disturbance in AMS, this study pivots towards a groundbreaking structural solution.

The paper presents a novel AMS equipped with a 4-DOF anthropomorphic tendon-driven manipulator. The design process involved the utilization of TAT structures for the elbow and wrist joints, and the application of finite element analysis, topology, and lattice optimization to select materials and achieve weight reduction. The kinematics were established based on a 7-DOF manipulator with virtual coupling joints, and theoretical models were developed to analyze joint torques and stiffness.By leveraging the remote actuation of tendon-driven systems and through strategic actuator distribution combined with topology and lattice optimization, we achieved a moving component weight of 0.818 kg and a base weight of 1.899 kg. This design effectively minimizes coupling disturbance in the AMS, as validated through our simulation environment.

In future endeavors, considering ATDM's adaptability for integrated waterproofing and electromagnetic interference shielding, we plan to deploy it for aerial operations in challenging environments.


\section*{Declaration of Competing Interest}
The authors declare that they have no known competing financial interests or personal relationships that could have appeared to influence the work reported in this paper.
\section*{Acknowledgment}
This work was supported partially by the National Natural Science Foundation of China (Grant No. 62273122), and Heilongjiang Natural Science Foundation (Grant No. YQ2023F009).

\section*{Data Availability}
Data will be made available on request.

\appendix
\section{Antiparallelogram Analysis}
\label{appendix:anti}

\begin{figure}[h!]
    \centering
    \subfigure[]{
        \includegraphics[width=0.23\textwidth]{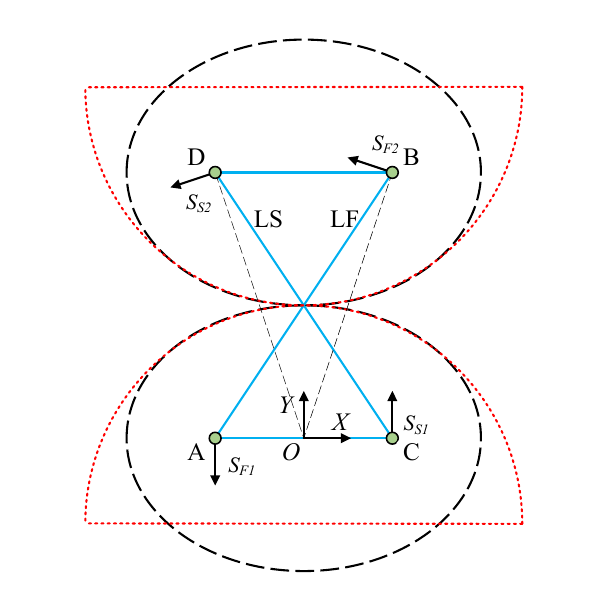}
        \label{fig:antiparallelogram a}
    }\hfill
    \subfigure[]{
        \includegraphics[width=0.23\textwidth]{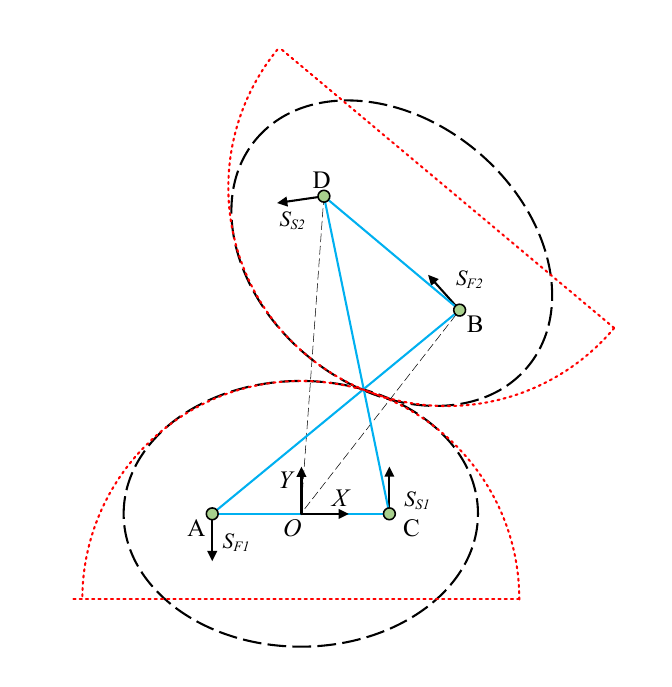}
        \label{fig:antiparallelogram b}
    }\\

    \subfigure[]{
        \includegraphics[width=0.23\textwidth]{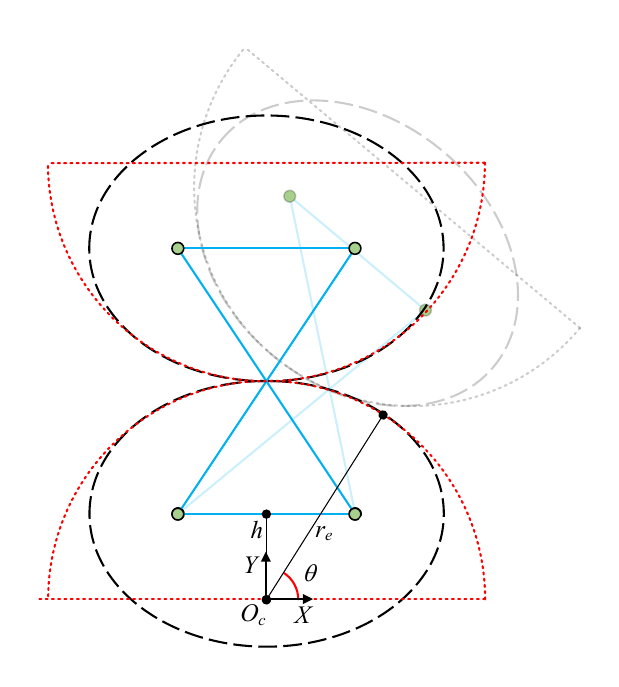}
        \label{fig:antiparallelogram c}
    }\hfill
    \subfigure[]{
        \includegraphics[width=0.23\textwidth]{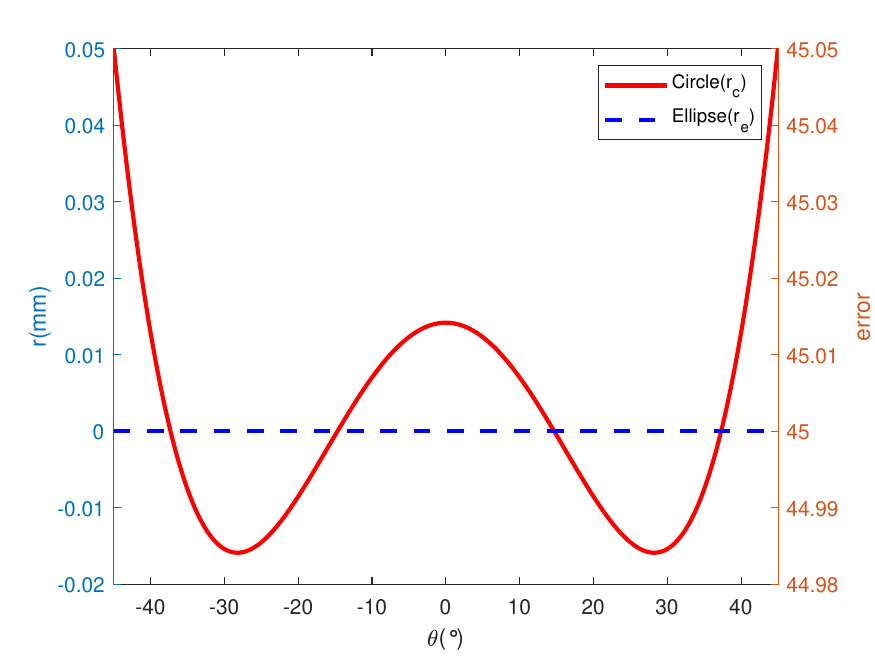}
        \label{fig:antiparallelogram d}
    }
    \caption{anti-parallelogram mechanism.(a)Anti-parallelogram mechanism (initial state).(b)Anti-parallelogram mechanism (Deflection state).(c)Approximation between circular rolling motion and elliptical rolling motion.(d)The error curve between ellipse and approximated circle}
    \label{figure:antiparallelogram}
\end{figure}

For planar motion, the antiparallelogram linkage can be designed such that the motion between the fixed platform and the moving platform resembles pure rolling on an approximate circular surface. The error of this is less than the machining error. Its fundamental principle is shown in Fig. (\ref{figure:antiparallelogram}).

The degree of freedom of the antiparallelogram mechanism was calculated using the screw theory. As illustrated, establish coordinate system $O_{A-XY}$ at the midpoint of AC. Assume the coordinates of B and C are $(a,b)$ and $(c,d)$ respectively. Let the two branches of the antiparallelogram mechanism be LF and LS. The four revolute pairs on these two branches are $\bm{S}_{F1}$, $\bm{S}_{F2}$, $\bm{S}_{S1}$, and $\bm{S}_{S2}$. The motion twist systems of the two branches are
{
\small
\setlength{\jot}{-2pt}
\def\arraystretch{0.7}
\begin{align}
&\left\{ {\begin{array}{*{20}{l}}
{{\bm{S}_{F1}} = \left( {\begin{array}{*{20}{l}}
0&0&{1;}&0&{ - w}&0
\end{array}} \right)}\\
{{\bm{S}_{F2}} = \left( {\begin{array}{*{20}{l}}
0&0&{1;}&{ - b}&a&0
\end{array}} \right)}
\end{array}} \right.{\rm{  }}\\
&{\rm{  }}\left\{ {\begin{array}{*{20}{l}}
{{\bm{S}_{S1}} = \left( {\begin{array}{*{20}{l}}
0&0&{1;}&0&w&0
\end{array}} \right)}\\
{{\bm{S}_{S2}} = \left( {\begin{array}{*{20}{l}}
0&0&{1;}&{ - d}&c&0
\end{array}} \right)}
\end{array}} \right.
\end{align}
}

The reciprocal twist systems (constraint twist systems) for each are
{
\footnotesize
\setlength{\jot}{-2pt}
\def\arraystretch{0.7}
\begin{align}
&\left\{
\begin{aligned}
\bm{S}_{LF1}^r &= \left( \begin{array}{cccccc} 0&0&1;&0&0&0 \end{array} \right) \\
\bm{S}_{LF2}^r &= \left( \begin{array}{cccccc} 0&0&0;&1&0&0 \end{array} \right) \\
\bm{S}_{LF3}^r &= \left( \begin{array}{cccccc} 0&0&0;&0&1&0 \end{array} \right) \\
\bm{S}_{LF4}^r &= \left( \begin{array}{cccccc} \frac{(a + w)}{b}&1&0;&0&0&w \end{array} \right)
\label{equation:constraint twist systems1}
\end{aligned}
\right.
\\
&\left\{
\begin{aligned}
\bm{S}_{LS1}^r &= \left( \begin{array}{cccccc} 0&0&1;&0&0&0 \end{array} \right) \\
\bm{S}_{LS2}^r &= \left( \begin{array}{cccccc} 0&0&0;&1&0&0 \end{array} \right) \\
\bm{S}_{LS3}^r &= \left( \begin{array}{cccccc} 0&0&0;&0&1&0 \end{array} \right) \\
\bm{S}_{LS4}^r &= \left( \begin{array}{cccccc} \frac{(c - w)}{d}&1&0;&0&0&{-w} \end{array} \right)
\label{equation:constraint twist systems2}
\end{aligned}
\right.
\end{align}
}
The linear sum of Equ.(\ref{equation:constraint twist systems1}) and Equ.(\ref{equation:constraint twist systems2}) is
{
\small
\setlength{\jot}{-2pt}
\def\arraystretch{0.7}
\begin{equation}
\left\{
\begin{aligned}
\bm{S}_{1}^r &= \left( \begin{array}{cccccc} 0&0&1;&0&0&0 \end{array} \right), \\
\bm{S}_{2}^r &= \left( \begin{array}{cccccc} 0&0&0;&1&0&0 \end{array} \right), \\
\bm{S}_{3}^r &= \left( \begin{array}{cccccc} 0&0&0;&0&1&0 \end{array} \right), \\
\bm{S}_{4}^r &= \left( \begin{array}{cccccc} \frac{(a + w)}{b}&1&0;&0&0&w \end{array} \right), \\
\bm{S}_{5}^r &= \left( \begin{array}{cccccc} \frac{(c - w)}{d}&1&0;&0&0&{-w} \end{array} \right)
\end{aligned}
\right.
\label{equation:anti_constraint_twist_systems}
\end{equation}
}

The motion twist system of the moving platform can be determined by finding the reciprocal twist system of Equ. (\ref{equation:anti_constraint_twist_systems}) as
{
\small
\setlength{\jot}{-2pt}
\def\arraystretch{0.7}
\begin{equation}
{\bm{S}^M} = \left( {\begin{array}{*{20}{l}}
0&0&{1;}&{\frac{{2wbd}}{{bc - wb - ad - wd}}}&{\frac{{{w^2}b - wbc - wad - {w^2}d}}{{bc - wb - ad - wd}}}&0
\end{array}} \right)
\end{equation}
}
Clearly, the antiparallelogram mechanism is a 1-DOF joint. When the link lengths are set such that AB=CD=$l$ and AC=BD=$2w$, the trajectory of the intersection point P can be described as an ellipse with A and C as its focus and $l$ as its major axis,which can be represented as
\begin{equation}
\frac{x^2}{(l/2)^2} + \frac{(y-h)^2}{(l/2)^2 - w^2} = 1
\label{equation:ellipse_of_anti}
\end{equation}

By satisfying the link length constraints, substituting the dual part of the motion twist of the moving platform into Equ.(\ref{equation:ellipse_of_anti}) satisfies the equation. When the parameters of the ellipse are chosen appropriately, over a certain range of motion, the elliptical arc can approximate a segment of a circular arc. To better compare the error during the rotation process, a polar coordinate system $O_{C-XY}$ is established at the center of the desired circle with the vertical direction as the polar axis, as shown in Fig.(\ref{fig:antiparallelogram c}). The polar radius of the ellipse can be expressed as:
\begin{equation}
   r_e = \frac{h \cos(\theta) \pm \sqrt{h^2 \cos^2(\theta) - 4 \left( \frac{\cos^2(\theta)}{l^2 - 4w^2} + \frac{\sin^2(\theta)}{l^2} \right) \left( \frac{4h^2}{l^2 - 4w^2} - 1 \right)}}{4 \left( \frac{\cos^2(\theta)}{l^2 - 4w^2} + \frac{\sin^2(\theta)}{l^2} \right)}
\end{equation}

Aim to use the segment of the ellipse where $-\pi/4 < \theta < \pi/4$ to approximate a segment of a circle with radius $r_c$, we calculate this using the following constraint optimization approach:

\begin{equation}
    \min (\int_{ - \frac{\pi }{4}}^{\frac{\pi }{4}} {\left| {{r_e}\left( \theta  \right) - {r_c}\left( \theta  \right)} \right|} d\theta ), \quad s.t \quad {r_c} = 40;w = 12;
\end{equation}

Upon calculation, with the ellipse's focal length set at $24$ and the desired circle's radius approximated at $40$, the values $h = 3.43$ and $l = 76.95$ yield the minimum approximation error. As shown in Fig.(\ref{fig:antiparallelogram d}), the fitting error across the entire motion space is less than 0.05mm, which is below the machining error. Due to the characteristic of pure rolling, the angular rotation range between the fixed platform and the moving platform is:$[ -\frac{\pi}{2}, \frac{\pi}{2} ]$.

\section{Movement Capabilities of Quaternion Joint}
\label{appendix:quaternion}
The quaternion joint, as theoretically analyzed, is completely constrained. However, experimental studies related to it have showcased its superior performance when functioning as a 2-DOF mechanism.
\begin{figure}[htbp]
    \centering
    \subfigure[]{
        \includegraphics[width=0.15\textwidth]{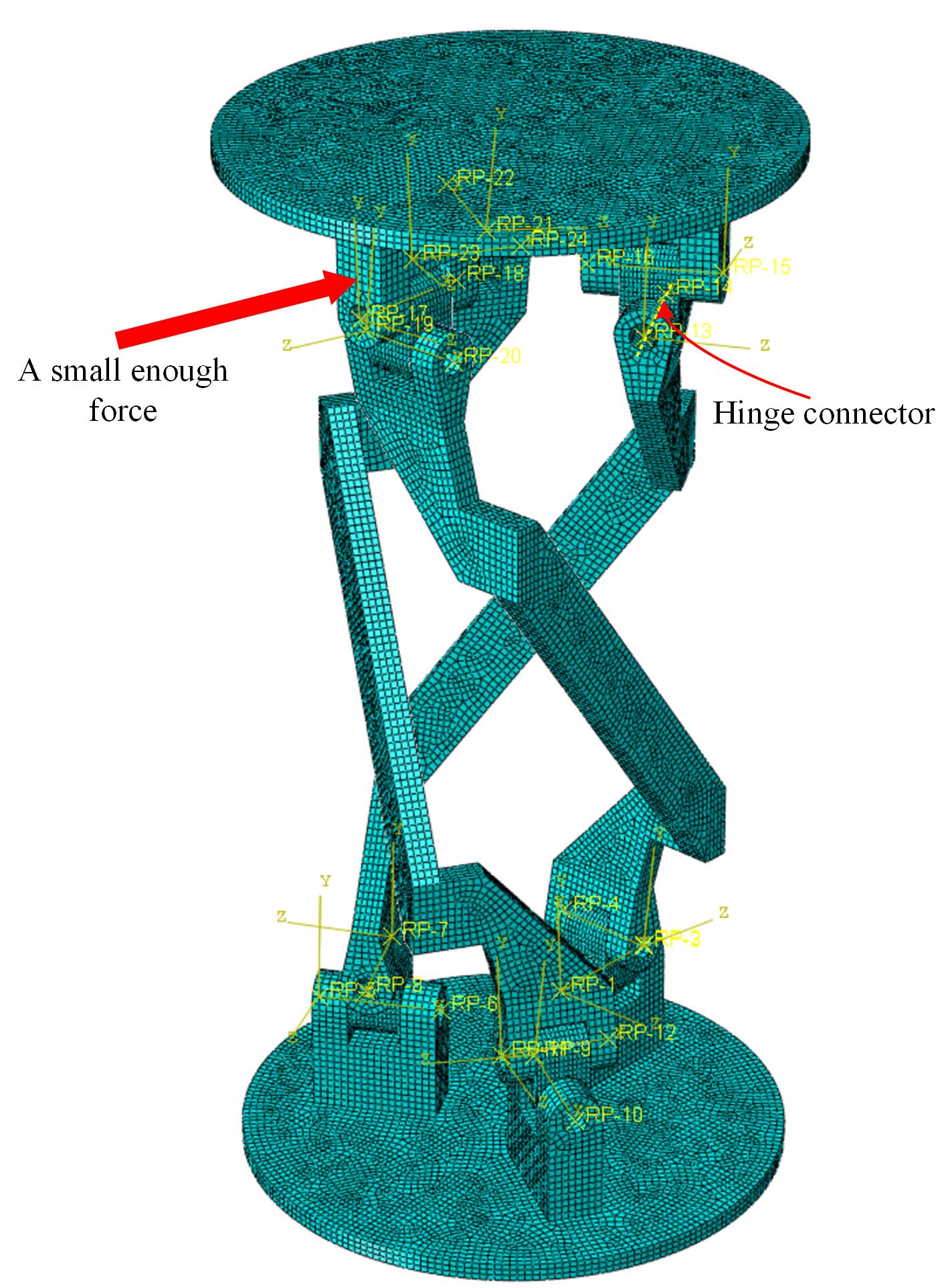}
        \label{appendix_quaternion_a}
    }\hfill
    \subfigure[]{
        \includegraphics[width=0.15\textwidth]{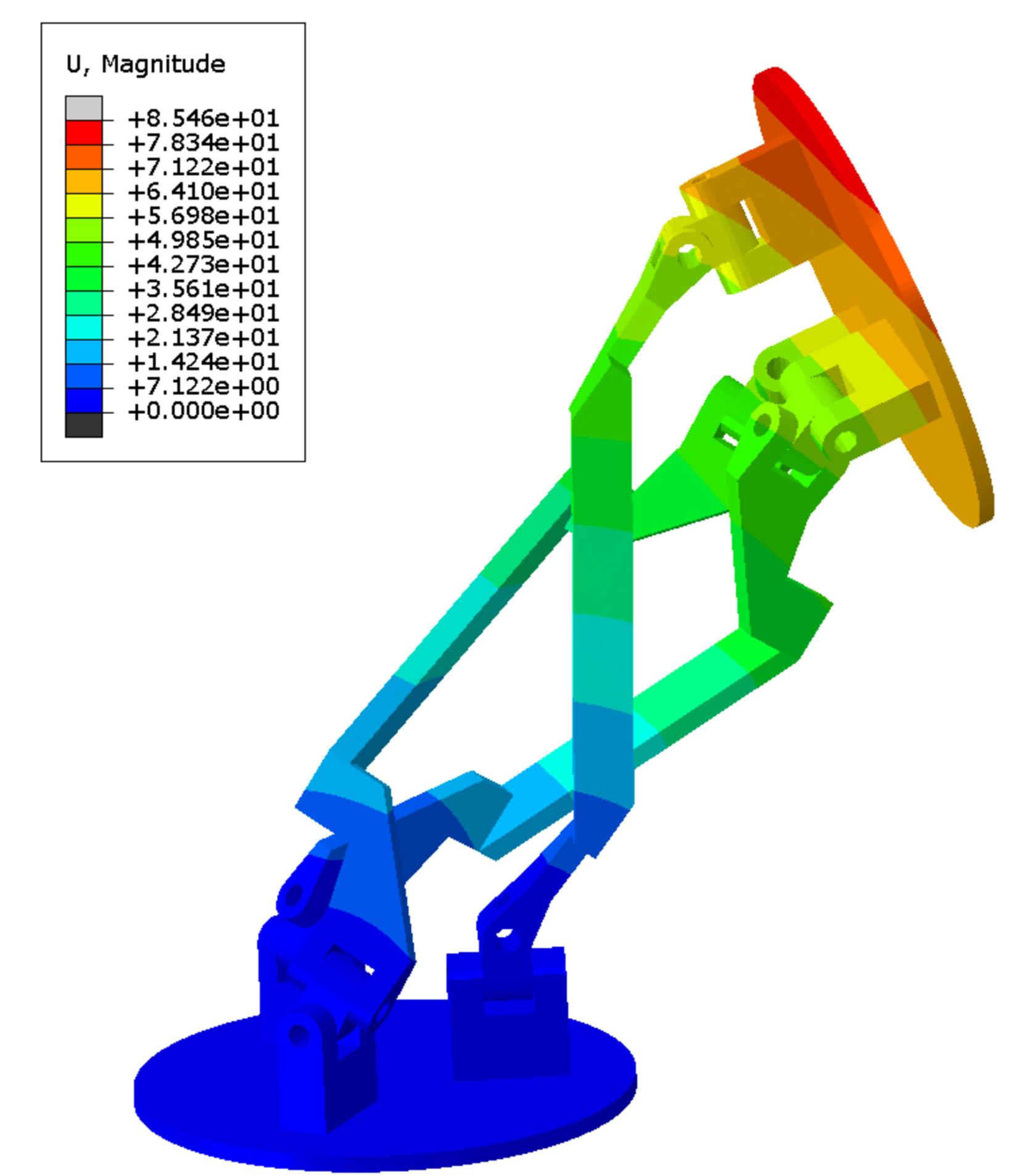}
        \label{appendix_quaternion_b}
    }\hfill
    \subfigure[]{
        \includegraphics[width=0.15\textwidth]{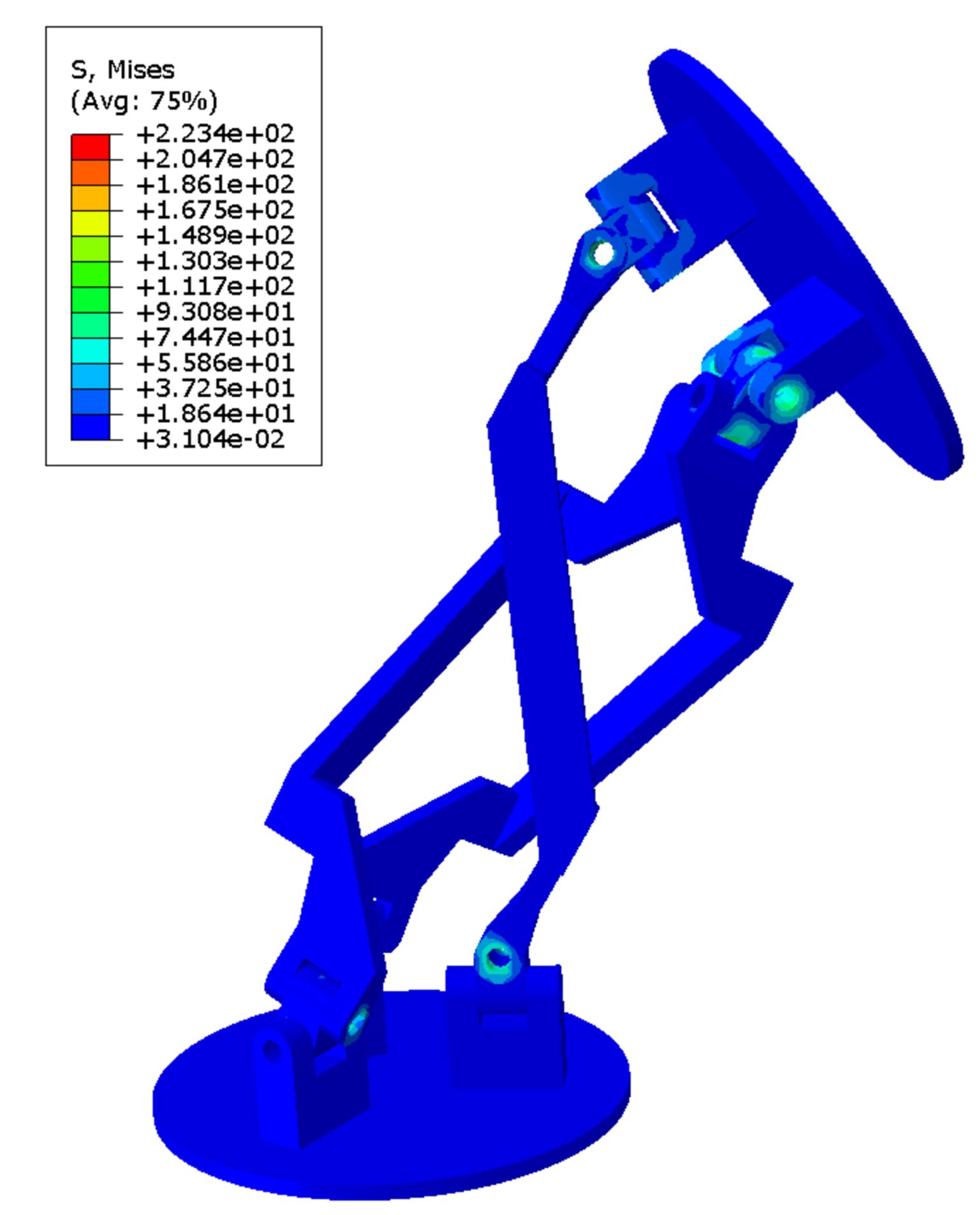}
        \label{appendix_quaternion_c}
    }
    \caption{Finite element analysis of quaternion joint.(a) Finite element model.(b)Displacement contour plot(mm).(c)Stress contour plot(Mpa)}
    \label{fig:quaternion}
\end{figure}

As depicted in Fig.~\ref{fig:quaternion}, a finite element model was developed in Abaqus (shown in Fig.~\ref{appendix_quaternion_a}). In this model, each pin in the structure was designated as a connector, with the connection category set to "hinge". Aluminum alloy 7075 was chosen as the material, and a general contact interaction was established. A small enough force of 0.01N was exerted on the moving platform. The explicit dynamics analysis results, as shown in Fig.~\ref{appendix_quaternion_b}, indicate the quaternion joint's capability for a wide range of motion. While there is stress concentration evident in the structure, as seen in Fig.~\ref{appendix_quaternion_c}, it remains unyielded. This attests to the mechanism's ability to move even under stress and micro-deformation conditions when subjected to minimal force.

\bibliographystyle{elsarticle-num} 
\begin{spacing}{1}
{\small
\setlength{\bibsep}{0pt} 
\bibliography{ATDM_cite}
}
\end{spacing}

\end{document}